\documentclass[journal]{IEEEtran}

\usepackage{cite}
\usepackage{amsmath,amssymb,amsfonts}
\usepackage{algorithmic}
\usepackage{graphicx}
\usepackage{textcomp}
\usepackage{algorithm,algorithmic}
\usepackage{hyperref}
\usepackage{makecell}
\usepackage{multicol}
\usepackage{multirow}
\usepackage{soul, color}
\usepackage[table]{xcolor}  
\usepackage{enumitem}
\usepackage{comment}
\usepackage{tikz}
\usetikzlibrary{arrows.meta, shapes, positioning, calc}

\ifCLASSINFOpdf
\else
\fi
\begin{document}
%
\title{Dexterous Manipulation through Imitation Learning: A Survey}
%
%
%

\author{Shan An,~\IEEEmembership{Senior Member,~IEEE,} Ziyu Meng, Chao Tang, Yuning Zhou, Tengyu Liu, Fangqiang Ding,\\
Shufang Zhang, Yao Mu,~\IEEEmembership{Member,~IEEE,}
Ran Song,~\IEEEmembership{Senior Member,~IEEE,} 
        Wei Zhang,~\IEEEmembership{Senior Member,~IEEE,} \\ Zeng-Guang Hou,~\IEEEmembership{Fellow,~IEEE}, 
        and Hong Zhang,~\IEEEmembership{Fellow,~IEEE}
\thanks{
The authors gratefully acknowledge the support of the National Key Research and Development Program of China (Grant No. 2023YFC3603601), and the National Natural Science Foundation of China (Grant No. U22A2057).
(Corresponding author: Ran Song)
}
\thanks{Shan An is with the Tianjin Key Laboratory of Intelligent Unmanned Swarm Technology and System, the School of Electrical and Information Engineering, Tianjin University, Tianjin 300072, China.
}
\thanks{Ziyu Meng is with the School of Control Science and Engineering, Shandong University, Jinan 250061, China, also with the State Key Laboratory of General Artificial Intelligence, Beijing 100086, China.}
\thanks{Chao Tang is with the Division of Robotics, Perception and Learning at KTH Royal Institute of Technology, Stockholm 11428, Sweden.}
\thanks{Ran Song and Wei Zhang are with the School of Control Science and Engineering, Shandong University, Jinan 250061, China. (e-mail: ransong@sdu.edu.cn).}
\thanks{Yuning Zhou is with the Department of Mechanical and Process Engineering, ETH Zurich, 8092 Zurich, Switzerland.}
\thanks{Tengyu Liu is with the State Key Laboratory of General Artificial Intelligence, Beijing 100086, China.}
\thanks{Fangqiang Ding is with the Department of Mechanical Engineering at the Massachusetts Institute of Technology (MIT), MA 02118, USA.}
\thanks{Shufang Zhang is with the School of Electrical and Information Engineering, Tianjin University, Tianjin 300072, China.}
\thanks{Yao Mu is with the School of Computer Science, Shanghai Jiao Tong University, Shanghai 200240, China.}
\thanks{Zeng-Guang Hou is with the State Key Laboratory of Multimodal Artificial Intelligence Systems, Institute of Automation, Chinese Academy of Sciences, Beijing 100190, China, also with the School of Artificial Intelligence, University of Chinese Academy of Sciences, Beijing 100049, China, and also with CASIA-MUST Joint Laboratory of Intelligence Science and Technology, Institute of Systems Engineering, Macau University of Science and Technology, Macao 999078, China.}
\thanks{Hong Zhang is with the Department of Electronic and Electrical Engineering, Southern University of Science and Technology, Shenzhen 518055, China.}
}


%
%

\markboth{Journal of \LaTeX\ Class Files,~Vol.~14, No.~8, Dec.~2025}%
{Shell \MakeLowercase{\textit{et al.}}: Bare Demo of IEEEtran.cls for IEEE Journals}
%



\maketitle


\begin{abstract}


Dexterous manipulation, which refers to the ability of a robotic hand or multi-fingered end-effector to skillfully control, reorient, and manipulate objects through precise, coordinated finger movements and adaptive force modulation, enables complex interactions similar to human hand dexterity. With recent advances in robotics and machine learning, there is a growing demand for these systems to operate in complex and unstructured environments. Traditional model-based approaches struggle to generalize across tasks and object variations due to the high dimensionality and complex contact dynamics of dexterous manipulation. Although model-free methods such as reinforcement learning (RL) show promise, they require extensive training, large-scale interaction data, and carefully designed rewards for stability and effectiveness. Imitation learning (IL) offers an alternative by allowing robots to acquire dexterous manipulation skills directly from expert demonstrations, capturing fine-grained coordination and contact dynamics while bypassing the need for explicit modeling and large-scale trial-and-error. This survey provides an overview of dexterous manipulation methods based on imitation learning, details recent advances, and addresses key challenges in the field. Additionally, it explores potential research directions to enhance IL-driven dexterous manipulation. Our goal is to offer researchers and practitioners a comprehensive introduction to this rapidly evolving domain.

\end{abstract}

\def\abstractname{Note to Practitioners}
\begin{abstract}
This work explores the intersection of IL and dexterous manipulation. With promising applications in manufacturing, healthcare, and home robotics, IL-based approaches allow robots to handle delicate objects, execute precise tasks, and operate in diverse, unstructured environments. However, key challenges remain—particularly in collecting high-quality demonstrations and enabling generalization from limited data, both of which are critical for real-world deployment. This survey aims to serve as a practical and accessible guide for practitioners seeking to apply IL-based methods to develop more capable and adaptable robotic systems in real-world scenarios.
\end{abstract}

\begin{IEEEkeywords}
Dexterous Manipulation, Imitation Learning, End Effector, Teleoperation
\end{IEEEkeywords}

\vspace{1cm}


%
\IEEEpeerreviewmaketitle



\section{Introduction}
%
%
%
%

\begin{figure}[t]
    \centering
    \includegraphics[width=1.0\linewidth]{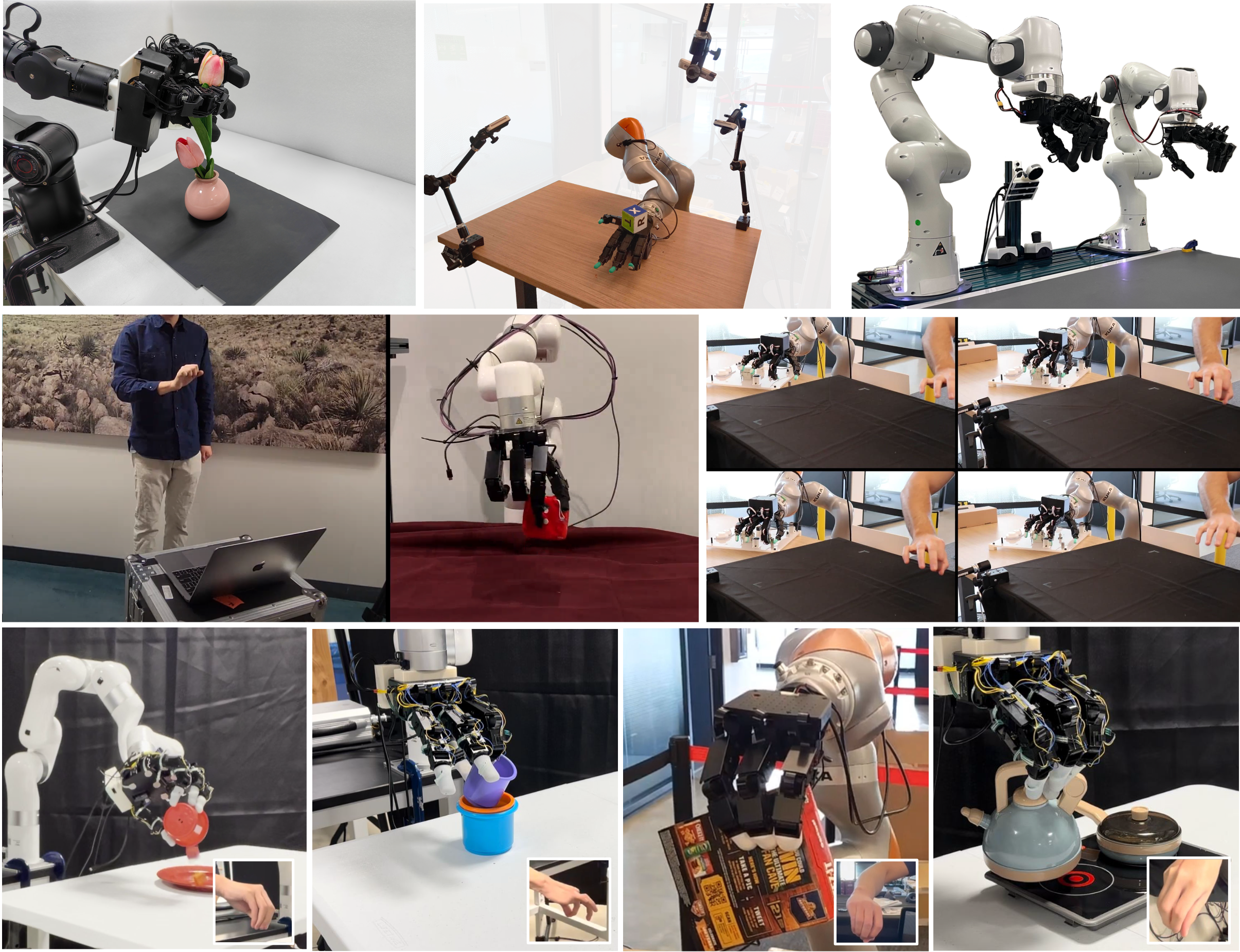}
    \vspace{-0.3cm} 
    \caption{Examples of dexterous manipulation in the real world. Row 1: Customized dexterous manipulation platform, Dextreme~\cite{handa2023dextreme}, DexCap~\cite{wang2024dex},
    Row 2: Robotic telekinesis~\cite{sivakumar2022robotic}, Dexpilot~\cite{handa2020dexpilot},
    Row 3: Anyteleop~\cite{qin2023anyteleop}.}
    \label{fig_1}
    \vspace{-0.2cm} 
\end{figure}

\IEEEPARstart{O}{ver} 
the past few decades, robotics has attracted intensive research interests, with dexterous manipulation emerging as a particularly popular focus. 
Dexterous manipulation aims to perform complex, precise, and flexible tasks (such as grasping an object, opening a drawer, and rotating a pencil) in various scenes with human-level dexterity using robotic hands or other end-effectors, as illustrated in Fig. \ref{fig_1}. This high-precision manipulation capability supports a broad spectrum of applications, including industrial manufacturing \cite{GONZALEZ2021283,8629046,liang2017using,rebelo2014bilateral}, space or underwater exploration, \cite{diftler2012robonaut,brantner2021controlling,barbieri2018design,gharaybeh2019telerobotic}, and medical care \cite{zhang2020microsurgical,guo2019scaled,pugin2011history,talamini2002robotic}.
Recently, the rapid development of imitation learning (IL) \cite{peng2018deepmimic,zhang2018deep}, which seeks to acquire knowledge by observing and mimicking behaviors of humans or other agents, has led to notable advancements in computer graphics and robotics. As an intuitive approach to equip robots with human prior knowledge, especially in the ability to interact with objects and understand scenes, IL has shown exceptional performance in enabling robots to perform tasks with human-like dexterity.

\begin{figure*}[t]
    \centering
    \includegraphics[width=\textwidth]{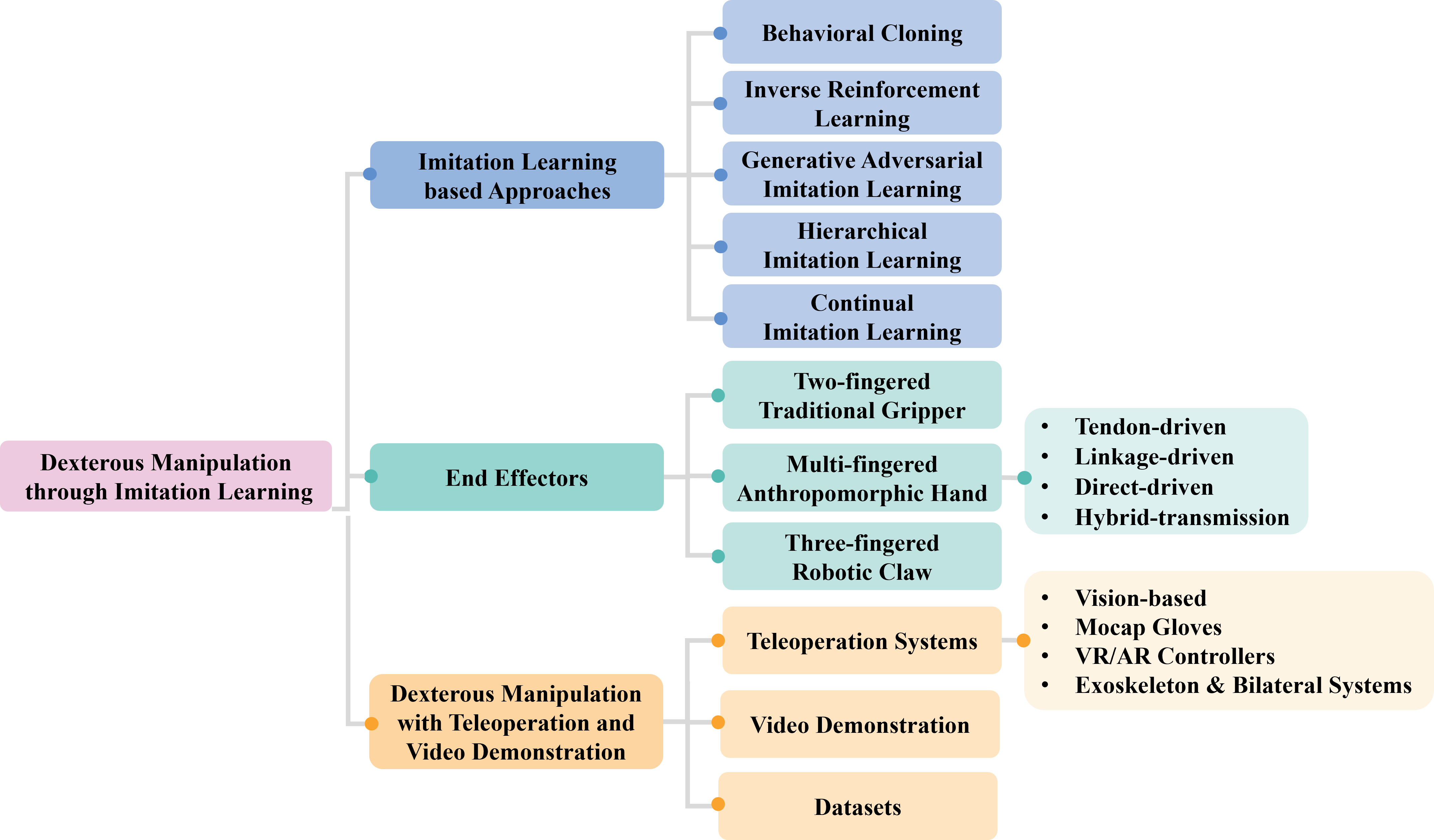}  
    \caption{Overview of imitation learning-based dexterous manipulation methods in this survey.}
    \label{fig:pdf_image}
\end{figure*}

Research on dexterous manipulation has received significant attention even before reinforcement learning (RL) was adopted to optimize behavior strategies through iterative interactions with the environment and reward-based feedback mechanisms. Traditional approaches encourage robots to acquire dexterous manipulation skills by modeling domain dynamics and applying optimal control methods. These approaches are theoretically sound but rely heavily on the fidelity of the world model.

However, when it comes to dexterous operations, such as assembling precision components or performing complex surgical procedures, a high degree of flexibility and multi-degree-of-freedom movement capabilities are required to execute complex, human-like tasks.
The successful execution of dexterous manipulation hinges upon intricate and highly precise mechanical design, such as multi-fingered robotic hand \cite{qin2022one} or anthropomorphic arms \cite{grannen2023stabilize}, as well as sophisticated control algorithms required to handle high-dimensional spaces \cite{grannen2023stabilize} and multi-contact dynamics \cite{zhu2023difflfd}.

Recently, the exploration of employing IL in the field of robotics has garnered significant attention from researchers. Without the need to craft complex world models and carefully designed reward functions, IL enables robots to learn tasks by observing and imitating expert demonstrations. 
This approach is intuitive, as the goal is for robots to substitute human labor by performing tasks like human experts. Specifically, the initial step involves collecting a dataset of expert demonstrations, which contains trajectories of manipulation tasks conducted by humans or well-trained agents. 
Robots use such trajectories as a reference to learn task behaviors. To ensure consistency, it is preferable to use identical robots during both the data collection and execution phases. However, this implementation does not facilitate data sharing among heterogeneous robotic systems. One solution is to map the trajectory of the original manipulator to the target robots, a process known as retargeting. Nonetheless, the process of humans operating robots for data collection remains time-consuming and labor-intensive in constructing large-scale datasets. To address this issue, researchers \cite{antotsiou2018task,li2019vision,sivakumar2022robotic} have adopted pose estimation techniques from computer vision to develop mappings from human hands to robotic hands, effectively lowering the barriers of collecting demonstration data. Additionally, dataset augmentation enhances the ability to generalize to new objects and scenes, contributing to the expansion of the dataset.

IL mimics expert behaviors analogous to supervised learning (SL) and often integrates with RL to address complex decision-making tasks.
Both IL and SL share similarities in learning from demonstrations or ground truth data. 
However, their objectives differ: SL aims to produce outputs identical to the ground truth in static scenarios, while IL focuses on task completion in dynamic environments, such as changes in target object position and environmental disturbances in manipulation tasks.
Such tasks usually involve sequential decision-making, where errors can accumulate over steps, leading to compounding errors and overall task failure. IL emphasizes task completion by adjusting and compensating for initial errors in subsequent decisions, thereby reducing their overall impact.
In dexterous manipulation tasks, RL and IL are often combined. 
This combination addresses the inefficiencies caused by the agent's large and complex action space, has a high degree of freedom, and makes pure RL exploration less effective. IL leverages expert demonstrations to offer straight guidance, thereby reducing exploration time and increasing efficiency. Additionally, reward functions for manipulation tasks are often challenging to design. Different tasks typically necessitate distinct reward functions. However, IL benefits from a relatively universal reward function to fit demonstration trajectories, necessitating only the provision of varied demonstration data, which ultimately improves learning efficiency and task success rates.

IL is particularly advantageous for dexterous manipulation tasks. This is because objects involved in dexterous manipulation are typically designed for human use, and thus, robots subject to these tasks are likely to have structures similar to humans or parts of the human body, such as humanoid robots, dual-arm manipulators, or dexterous hands. They usually require precise control, coordination, and adaptability, attributes that are challenging to achieve through traditional methods.
Specifically, this learning paradigm includes various branches such as behavior cloning \cite{shi2023waypoint}, \cite{florence2022implicit}, hybrid approaches (the combination of RL and IL) \cite{rajeswaran2018learning},  \cite{ding2019goal}, 
hierarchical IL \cite{mandlekar2020learning}, \cite{belkhale2023hydra}, and others, each contributing unique advantages to the learning process.

Since the intersection of IL and dexterous manipulation represents a frontier in robotics research. In the past decade, several works including DAPG \cite{rajeswaran2018learning}, which combined deep reinforcement learning(DRL) with human demonstrations to solve high-dimensional dexterous manipulation tasks; Implicit Behavioral Cloning \cite{florence2022implicit} which focused on improving robot policy learning from a mathematical perspective; Hiveformer \cite{guhur2023instruction}, which explored creating multimodal interactive agents; Diffusion Policy \cite{chi2023diffusion} leveraged recent advancement in generative models to achieve better performance in manipulation tasks, have been proposed and significantly expanded the boundaries of what is achievable in robotic dexterous manipulation. However, despite the recent considerable progress in this field, numerous challenges remain. Data collection for IL is labor-intensive and time-consuming \cite{zeng2021transporter}, \cite{wang2024dex}. The acquisition of generalization ability from learned behaviors to new tasks and varying environments is also non-trivial \cite{haldar2023teach}. Additionally, real-time control and sim-to-real transfer, where robots trained in the simulation must perform effectively in the real world, both hinder the application \cite{zare2023survey}. Addressing these challenges requires a concerted effort to develop more efficient data collection methods, improve learning algorithms, and enhance the physical capabilities of robotic systems.

Building on significant advancements in the intersection of IL and dexterous manipulation, several recent surveys have explored key aspects of this rapidly evolving field. Zare et al.  \cite{zare2024survey} provided an extensive overview of the applications, progress, and challenges of imitation learning in both robotics and artificial intelligence. Arora and Doshi \cite{ARORA2021103500} focused on inverse reinforcement learning (IRL), particularly the process of inferring reward functions from observed behavior. Han et al. \cite{s23073762} delved into deep reinforcement learning (DRL), reviewing recent methods designed to optimize DRL algorithms for real-world applications. Li et al. \cite{li2025developments} offered a comprehensive review of data collection methods and skill learning frameworks for robotic dexterous manipulation, emphasizing the key challenges in this area. Pitkevich and Makarov \cite{pitkevich2024survey} addressed the Sim-to-Real transfer problem in DRL, discussing the deployment of robots trained in simulation to real-world environments. Welte and Rayyes \cite{welte2025interactive} examined the potential of interactive imitation learning for humanoid robots, with a focus on how these methods can be transplanted for dexterous tasks. Finally, Tsuji et al. \cite{tsuji2025survey} provided a systematic review of imitation learning techniques for contact-rich tasks, discussing the unique challenges and emerging trends in this field.
In contrast to these existing works, which primarily focus on specific aspects of imitation learning or dexterous manipulation, this survey aims to provide a comprehensive overview of IL-based dexterous manipulation approaches.

The rest of this article is organized as follows: Section II presents an introduction to dexterous manipulation and IL in detail. Subsequently, we discuss the state-of-the-art IL-based dexterous manipulation techniques and highlight notable achievements in this field in Section III. Section IV discusses end-effectors for dexterous manipulation. Moreover, we discuss teleoperation systems and datasets in Section V. We summarize existing challenges and propose future directions for research in this rapidly evolving field in Section VI. Finally, conclusions are made in Section VII.
An overview of this survey is shown in Fig.~\ref{fig:pdf_image}. By synthesizing the existing body of knowledge, this survey aspires to serve as a valuable resource for researchers and practitioners seeking to advance the capabilities of robotic systems through the synergy of IL and dexterous manipulation.

\section{Overview of Imitation Learning based Dexterous Manipulation}

\subsection{Dexterous Manipulation}
In the field of robotics, dexterous manipulation \cite{salisbury1982articulated, Mason1985RobotHA, bai2014dex, liu2025fusion} refers to the capability of robotic systems to execute intricate and precise tasks. These tasks often employ grippers or dexterous hands to grasp, maneuver, and manipulate objects \cite{okamura2000overview}. Characterized by high degrees of freedom and fine motor skills, dexterous manipulation extends beyond simple pick-and-place operations to include activities such as tool use, object reorientation, and complex assembly tasks. Achieving such manipulation commonly involves using sophisticated end-effectors designed to emulate the versatility and finesse of human hands, such as multi-fingered hands or anthropomorphic robotic arms.

Dexterous manipulation poses several significant challenges, including precise control, high-dimensional motion planning, and real-time adaptability to dynamic environments \cite{yu2022dexterous}. The intricacies of these tasks demand robust mechanical design and advanced control algorithms capable of handling the complexities of multi-contact interactions and the variability inherent in real-world scenarios. Traditional model-based methods \cite{buss1996dextrous, mordatch2012contact} have become inadequate for robots performing complex tasks due to the increasing complexity of manipulation tasks. Consequently, extensive research has been dedicated to learning-based approaches, with RL emerging as an effective method. Various works have exploited RL for robots to learn dexterous policies \cite{popov2017data, williams1992simple, schulman2017proximal, dai2022analysing, xu2023dexterous, yin2023rot}. However, pure RL has several inherent drawbacks. RL algorithms often struggle with exploring high-dimensional action spaces efficiently in dexterous manipulation tasks \cite{lillicrap2015continuous}. Additionally, designing reasonable reward functions is challenging; flawed reward functions can affect exploration and learning speed, leading to inferior performance. Recently, advancements in IL have opened new avenues for addressing these challenges. 

\subsection{Imitation Learning}
The main purpose of IL is to enable agents to learn and perform behaviors by imitating expert demonstrations \cite{zare2023survey}. 
In contrast, pure RL requires carefully designed reward functions and is particularly effective in scenarios where the desired behavior is difficult to describe in algorithms but can be easily demonstrated. IL employs these expert demonstrations to guide the learning process of agents by establishing the correlation between observed states and corresponding actions. Through IL, agents can transcend merely replicating basic and predefined behaviors within controlled and constrained environments, enabling them to autonomously execute optimal actions in complex, unstructured environments \cite{ravichandar2020recent}. Consequently, IL significantly alleviates the burden on experts, facilitating efficient skill transfer.

Methodologies of IL can be broadly categorized into several sub-classes, including behavior cloning \cite{bain1995framework}, inverse reinforcement learning (IRL) \cite{ng2000algorithms}, and generative adversarial imitation learning (GAIL) \cite{ho2016gener}. 
Behavior cloning directly maps observed actions to the agent’s actions through SL techniques. Conversely, IRL aims to deduce the underlying reward structure that motivates the demonstrator’s behavior, allowing the agent to optimize its actions accordingly. GAIL employs adversarial training techniques to improve the imitation policy by distinguishing between expert and agent actions, thus refining the agent's ability to replicate the desired behavior accurately.

\color{black}
\subsection{Connections to Human Motor Learning}
IL is fundamentally motivated by the way humans acquire motor skills through observation, demonstration, and practice. From a cognitive perspective, Bandura’s social learning theory~\cite{bandura1977social} emphasized that behavior can be learned by observing others, rather than solely through reinforcement or trial-and-error. This insight has inspired IL paradigms that rely on demonstration-based policy generation instead of environment-driven exploration.

At the neural level, the discovery of mirror neurons in primates provides biological evidence that action observation and execution may share overlapping representations~\cite{rizzolatti2004mirror}. These findings suggest that learning by watching others perform a task can activate internal motor plans, a mechanism analogous to behavioral cloning in robotics. Furthermore, models of human motor control, such as the internal model theory~\cite{wolpert1995internal} and optimal feedback control~\cite{todorov2002optimal}, offer insights into how the brain predicts, plans, and corrects motor behaviors. These principles have influenced the design of IL algorithms that incorporate predictive models or closed-loop feedback refinement. For example, Nah et al.~\cite{nah2025modular} presented a modular robot control framework using Dynamic Movement Primitives (DMP) and Elementary Dynamic Actions (EDA), employing dynamical systems to predict and correct motor behaviors in a manner similar to internal models in the brain.

In summary, the foundations of IL are deeply rooted in cognitive and neuroscientific principles, from observational learning and mirror neuron systems to predictive motor control and hierarchical skill organization. These biological insights continue to shape IL algorithms, guiding the development of methods that mimic the way humans learn and adapt motor behaviors.

\subsection{Long-horizon Dexterous Manipulation}

Long-horizon dexterous manipulation involves executing complex tasks that require both fine-grained control and long-term planning. 
These tasks typically span long time horizons and require coherent, efficient execution of sequential actions.
Without an appropriate structure, such tasks can overwhelm agents, complicating both planning and execution.
Task decomposition addresses this challenge by enabling agents to focus on different task components at various levels of abstraction, thereby facilitating the management of long-horizon manipulation \cite{chen2025deco}. In the context of IL, task decomposition is commonly realized through temporal abstraction, often implemented via techniques such as option discovery~\cite{achiam2018variational} or hierarchical policy structures~\cite{sun2024hierarchical,wang2023hierarchical}.

Temporal abstraction enables an agent to make decisions through temporally extended action sequences, thereby accelerating learning and simplifying complex tasks. Option discovery refers to the automated process of identifying and constructing these temporal abstractions, known as options, that comprise an internal low-level policy, an initiation set, and a termination condition. The options framework \cite{sutton1999between, precup2000temporal} generalizes the traditional action concept by treating options as temporally extended actions, allowing planning over multiple time scales.
The option-critic architecture~\cite{bacon2017option} and its variants are widely adopted to automatically extract options from demonstration data, which are then scheduled by high-level policies to effectively organize the temporal structure of tasks. For example, DDCO~\cite{krishnan2017ddco} autonomously discovers and learns hierarchical options within continuous action spaces from expert demonstrations, enabling unsupervised task decomposition and hierarchical IL for robotic control.

Hierarchical RL is often integrated with IL to tackle long-horizon tasks and complex policy learning \cite{li2021hierarchical}. The fundamental principle involves leveraging human demonstrations to decompose tasks along temporal or functional dimensions, which reduces the policy search space and enhances both learning efficiency and generalization.  For example, SRT-H~\cite{kim2025srt} proposed a hierarchical language-conditioned IL framework that integrates high-level instruction parsing with low-level dexterous control to allow the execution of end-to-end autonomous surgical tasks. 

\subsection{Multi-agent or Collaborative Dexterous Manipulation}
Multi-agent or collaborative dexterous manipulation refers to the system-level problem in which two or more agents endowed with multi-fingered dexterous manipulation capabilities, such as robot arms, dexterous hands, or mobile manipulators, jointly accomplish a high-dimensional, long-horizon, contact-rich task. The scenario emphasizes:
\begin{itemize}
    \item heterogeneous or homogeneous agents sharing workspace and task objectives;
    \item explicit or implicit communication (force, vision, language instructions) for action synchronization and load distribution;
    \item leveraging the redundant degrees of freedom of multiple arms/hands to realize complex operations unattainable by a single arm (e.g., bimanual valve turning, multi-human cooperative suturing).
\end{itemize}

To enable collaborative dexterous manipulation, Kim et al.~\cite{kim2025srt} introduced the SRT-H framework, which employs a hierarchical policy integrating high-level language instructions with low-level trajectory imitation. In a dual-arm da Vinci suturing task, the framework achieves precise bimanual coordination over long horizons, relying solely on human demonstrations without any reward supervision. Building on this line of research, BUDS~\cite{grannen2023stabilize} introduced a ``stabilize-then-act’’ role decomposition with a visual keypoint stabilization mechanism, attaining 76.9\% task success on complex bimanual dexterous tasks without reward signals or RL fine-tuning. Extending further, Bi-DexHands~\cite{chen2023bi} framed bimanual dexterous manipulation as a comprehensive suite of over 20 cooperative sub-tasks, providing the first large-scale benchmark for evaluating the scalability of multi-agent RL in physically coordinated manipulation. More recently, BiDexHD~\cite{zhou2024learning} advanced collaborative dexterous manipulation by integrating multi-task human demonstration data with teacher–student policy learning. This framework enables robots to efficiently acquire a wide range of cooperative bimanual skills across numerous auto-constructed tasks, substantially improving task completion rates and zero-shot generalization.

While task-level coordination addresses high-level planning, physical-level coordination must handle coupled, contact-rich dynamics inherent to dexterous manipulation. This challenge is addressed through unified model-based control. For example, Sleiman et al.~\cite{sleiman2021unified} proposed a whole-body MPC framework that jointly optimizes contact forces and centroidal dynamics in real time, enabling coherent coordination of locomotion and manipulation. Likewise, Yu et al.~\cite{yu2025generalizable} integrated global planning with adaptive MPC for dual-arm manipulation of deformable objects, capturing coupling through predictive dynamics. Beyond conventional hands, Zhong et al.~\cite{zhong2025paired} demonstrated that even micro-scale multi-agent coordination requires explicit handling of coupling dynamics, modeling nonlinear magnetic interactions between paired millirobots via data-driven inversion control and active disturbance rejection. These approaches collectively demonstrate that real-time compensation of coupled dynamics is essential for safe, responsive, and generalizable multi-agent IL policies.


In general, the core challenge of multi-agent or collaborative dexterous manipulation lies in learning generalizable cooperative policies within large-scale continuous control spaces while satisfying safety, real-time responsiveness, and consistency with human demonstrations, particularly under the imitation-learning paradigm.

\color{black}

\color{black}
\begin{table*}[t]
\centering
\scriptsize
\renewcommand{\arraystretch}{1.5}
\setlength{\tabcolsep}{5pt}
\caption{Comparison of different imitation learning approaches}
\begin{tabular}{m{3cm}|m{4cm}|m{3cm}|m{3cm}|m{3cm}}
\hline
\textbf{Approach} & \textbf{Key Characteristics} & \textbf{Pros} & \textbf{Cons} & \textbf{Key Applications} \\
\hline
\textbf{Behavioral Cloning} \cite{zhao2023learning,arunachalam2023dexterous,qin2022one, 8462901, zhang2018deep, osa2018algorithmic,ke2021grasping, mandlekar2020learning, zhao2023learning, florence2022implicit, shafiullah2022behavior, 9196935, chen2023diffusion, chi2023diffusion, ze2024d, ke20243d} & \begin{itemize}[leftmargin=3mm, rightmargin=0mm]
    \item Supervised learning paradigm.
    \item Mapping from states to actions.
    \item No reward signals or exploration.
\end{itemize} & \begin{itemize}[leftmargin=3mm, rightmargin=0mm]
    \item Simple and easy to implement.
    \item Data-efficient with large \newline demonstrations.
\end{itemize} & \begin{itemize}[leftmargin=3mm, rightmargin=0mm]
    \item Prone to distribution shift.
    \item Poor generalization to \newline unseen states.
\end{itemize} & \begin{itemize}[leftmargin=3mm]
    \item \textcolor{black}{Grasping and pick-and-place tasks.}
    \item \textcolor{black}{Short-horizon, structured tasks with sufficient data.}
\end{itemize} \\
\hline
\textbf{Inverse Reinforcement \newline Learning} \cite{9515637, 1603.00448, batzianoulis2021, millan2008, 2207.14299, app142311131, schulman2017proximal, 2412.11360, ARORA2021103500, s23073762, 10493015} & \begin{itemize}[leftmargin=3mm, rightmargin=0mm]
    \item Inferring expert’s reward function.
    \item Deriving policy by maximizing the \newline inferred reward.
\end{itemize} & \begin{itemize}[leftmargin=3mm, rightmargin=0mm]
    \item Generalization to new situations.
    \item Suitable for tasks with unknown \newline rewards.
\end{itemize} & \begin{itemize}[leftmargin=3mm, rightmargin=0mm]
    \item Computationally intensive.
    \item Non-unique reward solutions.
\end{itemize} & \begin{itemize}[leftmargin=3mm]
    \item \textcolor{black}{Complex tool-use tasks.}
    \item \textcolor{black}{Assembly tasks with implicit objectives.}
\end{itemize}\\
\hline
\textbf{Generative Adversarial \newline Imitation Learning} \cite{tangkaratt2019vild,  wang2021learning, zuo2021adversarial, antotsiou2021adversarial, liu2019hindsight, jiang2023mastering, wang2017robust, xiao2019wasserstein, arjovsky2017wasserstein, vahabpour2024diverse, tsurumine2022goal, shi2024ranking, zolna2021task, antotsiou2018task, tsurumine2019generative} & \begin{itemize}[leftmargin=3mm, rightmargin=0mm]
    \item Adversarial training between generator and discriminator.
    \item No explicit reward function.
\end{itemize} & \begin{itemize}[leftmargin=3mm, rightmargin=0mm]
    \item Good sample efficiency.
    \item Improved robustness.
\end{itemize} & \begin{itemize}[leftmargin=3mm, rightmargin=0mm]
    \item Training instability.
    \item Mode collapse and sensitivity \newline to hyperparameters.
\end{itemize} & \begin{itemize}[leftmargin=3mm]
    \item \textcolor{black}{Long-horizon dexterous tasks.}
    \item \textcolor{black}{In-hand manipulation with sparse demonstrations.}
    \end{itemize}\\
\hline
\textbf{Hierarchical Imitation \newline Learning} \cite{Kipf2018CompositionalIL, Xie2020DeepIL, 2409.16451, Xu2023XSkillCE, Wan2023LOTUSCI, wang2023mimicplay, 10610084} & \begin{itemize}[leftmargin=3mm, rightmargin=0mm]
    \item Two-level hierarchical policy.
    \item Decomposing tasks into sub-tasks \newline and primitives.
\end{itemize} & \begin{itemize}[leftmargin=3mm, rightmargin=0mm]
    \item Scalable to complex tasks.
    \item Modular and reusable \newline sub-policies.
\end{itemize} & \begin{itemize}[leftmargin=3mm, rightmargin=0mm]
    \item Requiring hierarchy design.
    \item Requiring training coordination.
\end{itemize} & \begin{itemize}[leftmargin=3mm]
    \item \textcolor{black}{Multi-step assembly and contact-rich tasks.}
    \item \textcolor{black}{Skill chaining and long-horizon manipulation.}
\end{itemize} \\
\hline
\textbf{Continual Imitation \newline Learning} \cite{Liang2024NeverEndingBA, Liu2023TAILTA, Wan2023LOTUSCI, Haldar2023PolyTaskLU, Gao2021CRILCR, Mete2024QueSTSS} & \begin{itemize}[leftmargin=3mm, rightmargin=0mm]
    \item Continual skill acquisition.
    \item Adapting previously learned behaviors.
\end{itemize} & \begin{itemize}[leftmargin=3mm, rightmargin=0mm]
    \item Flexible to evolving tasks.
    \item Reducing forgetting of old skills.
\end{itemize} & \begin{itemize}[leftmargin=3mm, rightmargin=0mm]
    \item Risk of catastrophic forgetting.
    \item Requiring ongoing expert input.
\end{itemize} & \begin{itemize}[leftmargin=3mm]
    \item \textcolor{black}{Lifelong learning for multi-task manipulation.}
    \item \textcolor{black}{Adapting to changing tools or objects.}
\end{itemize}\\
\hline
\end{tabular}
\label{tab:imitation_learning_comparison}
\end{table*}

\section{Imitation Learning based Dexterous Manipulation Approaches}
We categorize IL-based dexterous manipulation approaches into four categories: (1) Behavioral Cloning, (2) Inverse Reinforcement Learning, (3) Generative Adversarial Imitation Learning, and other extended frameworks, including 
(4) Hierarchical Imitation Learning and (5) Continual Imitation Learning. In the following subsections, we provide an overview of each category, followed by a detailed description and a summary of key research progress. Tab. \ref{tab:imitation_learning_comparison} presents a comparison between different IL approaches.

\subsection{Behavioral Cloning}
\subsubsection{Description} 
Behavioral Cloning (BC) refers to replicating expert behavior by learning directly from demonstrated state-action pairs. Specifically, BC is characterized by (1) a supervised learning paradigm and (2) a direct mapping from states to actions without relying on reward signals or exploration, as is typical in RL.


To formally define BC, we consider a set of $n$ demonstrations 
$\mathcal{D} = \{\tau_1, \dots, \tau_n\}$, where each demonstration $\tau_i$ is a sequence of state-action pairs of length $N_i$. Specifically, 
$\tau_i = \{(s_1, a_1), \dots, (s_{N_i}, a_{N_i})\}$, with states $s \in \mathcal{S}$ and actions $a \in \mathcal{A}$. $\mathcal{S}$ and $\mathcal{A}$ denote the state and action spaces, respectively. The objective of BC is to learn a policy $\pi: \mathcal{S} \rightarrow \mathcal{A}$ that imitates the expert behavior by minimizing the negative log-likelihood of the demonstrated actions. Formally, the objective function is:
\begin{equation}
\mathcal{L} (\pi) = - \mathbb{E}_{(s, a) \sim p_\mathcal{D}} \left[ \log \pi(a \mid s) \right].
\end{equation}

\subsubsection{Research Progress}
BC has achieved significant progress in dexterous manipulation \cite{zhao2023learning,arunachalam2023dexterous,10064325,10802784,qin2022one} and has demonstrated effective performance in relatively simple tasks, such as pushing \cite{8462901} and grasping \cite{zhang2018deep}. However, its applicability in dynamic environments and long-horizon tasks remains an active area of research.

The training data for BC models are usually collected from expert demonstrations tailored to specific tasks. Consequently, when the agent encounters states that are unseen during training, it may produce actions that deviate from the expert's behavior, leading to task failure. In sequential decision-making processes, even small deviations from expert actions at each step can accumulate over time, resulting in what is known as the ``compounding error'' problem. This issue is particularly pronounced in dexterous manipulation tasks\cite{osa2018algorithmic,ke2021grasping}, due to the high dimensionality of the action space and the strong dependency between task success and the consistency of the predicted action trajectory. To mitigate compounding errors in dexterous manipulation, Mandlekar et al.~\cite{mandlekar2020learning} proposed a hierarchical framework that segments demonstration trajectories at intersection points across different tasks and recombines them to synthesize trajectories for unseen tasks. Similarly, Zhao et al.~\cite{zhao2023learning} addressed the problem by considering the compatibility with high-dimensional visual observations. Instead of predicting actions step by step, they propose to predict entire action sequences, thereby reducing the effective decision horizon and alleviating compounding errors.

Another challenge in BC is its limited ability to model multimodal data, which is prevalent in human demonstrations collected from real-world environments. To overcome this limitation, several approaches have been proposed to model multi-modal action distributions. Florence et al.\cite{florence2022implicit} formulated BC as a conditional energy-based modeling problem for capturing multi-modal data distribution, albeit at the cost of increased computational overhead. Similarly, Shafiullah et al.\cite{shafiullah2022behavior} proposed to model the action distribution as a mixture of Gaussians. Their method leverages the Transformer architecture to efficiently utilize the history of previous observations and enables multi-modal action prediction through token-based outputs. Another promising direction is leveraging generative models to capture the inherent diversity of expert behaviors. Mandlekar et al.\cite{9196935} proposed using generative models for trajectory prediction, enabling selective imitation, though this approach relies on carefully curated, task-specific datasets. 

More recently, diffusion models have shown great potential in enhancing the robustness and generalization of BC methods. Chen et al.~\cite{chen2023diffusion} proposed a diffusion-augmented BC framework that models both conditional and joint probability distributions over expert demonstrations. Building on this idea, Chi et al.~\cite{chi2023diffusion} employed diffusion models as decision models to directly generate sequential actions conditioned on visual input and the robot's current state. Additionally, the 3D Diffusion Policy~\cite{ze2024d} leveraged 3D input representations to better capture scene spatial configurations. Similarly, the 3D Diffuser Actor~\cite{ke20243d} utilized full 3D scene representations by integrating RGB and depth information, along with language instructions, robot proprioception, and noise trajectories, through a 3D Relative Transformer framework.

\color{black}
Compared to variational autoencoders (VAEs)~\cite{kingma2019introduction} and generative adversarial networks (GANs)~\cite{goodfellow2020generative}, diffusion models offer several advantages for imitation learning, particularly in modeling complex, multimodal action distributions~\cite{chi2023diffusion,yan2025m}. 
Chen et al.~\cite{chen2023diffusion} proposed using diffusion models to augment BC and conducted a comprehensive comparison with other generative models, including VAEs, GANs, and EBMs (energy-based models)/implicit models, across different implementations in terms of model architecture, sampling/inference cost, and performance on various continuous control and manipulation tasks. Experimental results demonstrated that diffusion-augmented BC outperformed standard BC and several other generative model variants on multiple tasks.
\color{black}

\subsubsection{Discussion} In general, BC-based methods struggle with generalization and modeling multi-modal action distributions. To overcome these limitations, diffusion models have recently attracted increasing attention. They can be employed either as decision models that directly generate action sequences~\cite{chi2023diffusion} or as high-level strategy models that guide the action generation process~\cite{chen2023diffusion}. In both settings, diffusion models have shown promising performance and improved flexibility over conventional BC methods.

\subsection{Inverse Reinforcement Learning}
\subsubsection{Description}

Inverse Reinforcement Learning (IRL) inverts the conventional RL framework, which focuses on learning a policy to maximize a predefined reward function. Instead, IRL aims to infer the underlying reward function that best explains a set of expert demonstrations.


Formally, IRL estimates a reward function \( R(s, a) \) that best aligns with the demonstrated state-action pairs \( \mathcal{D} = \{\tau_1, \tau_2, \dots, \tau_N\} \), where \( \tau_i = \{(s_0, a_0), (s_1, a_1), \dots, (s_t, a_t) \} \). It is assumed that these demonstrations are generated by an expert following an optimal or near-optimal policy. The IRL problem is typically formulated within a finite Markov Decision Process, defined as \( \mathcal{M} = \langle \mathcal{S}, \mathcal{A}, T, R, \gamma \rangle \), where \( \mathcal{S} \) and \( \mathcal{A} \) are the state and action spaces, \( T(s'|s, a) \) is the state transition probability, \( R(s, a) \) is the reward function, and \( \gamma \in [0, 1] \) is the discount factor. IRL often represents the reward function as a linear combination of feature functions:
\begin{equation}
R(s_t, a_t) = w^\top \phi(s_t, a_t)
\end{equation}
where \( \phi(s, a) \) is a feature vector and \( w \) is a learnable weight vector. The expected feature counts under a policy \( \pi \) are defined as:
\begin{equation}
\mu_{\phi}(\pi) = \sum_{t=0}^\infty \gamma^t \psi^\pi(s_t) \phi(s_t, a_t)
\end{equation}
where \( \psi^\pi(s) \) denotes the state-action visitation frequency:
\begin{equation}
\psi^\pi(s) = \psi_0(s) + \gamma \sum_{s' \in S} T(s'|s, a) \psi^\pi(s').
\end{equation}

IRL is particularly advantageous in dexterous manipulation scenarios, where manually defining a reward function is often challenging or impractical. IRL has demonstrated effectiveness in various dexterous manipulation tasks, including dexterous grasping, assembly, and manipulation in dynamic and uncertain environments.


\subsubsection{Research Progress}
Recent studies have leveraged IRL frameworks to tackle complex dexterous manipulation tasks. Orbik et al.~\cite{9515637} first advanced IRL for dexterous manipulation by introducing reward normalization, task-specific feature masking, and random sample generation. These techniques effectively mitigate reward bias toward demonstrated actions and enhance learning stability in high-dimensional state-action spaces, leading to better generalization across unseen scenarios. Building upon the need for efficient learning in such high-dimensional settings, Generative Causal Imitation Learning~\cite{1603.00448} improved the sample efficiency of IRL by integrating maximum entropy modeling with adaptive sampling strategies. By leveraging nonlinear function approximation through neural networks, the proposed method enables expressive cost function learning while handling unknown system dynamics. To further incorporate user feedback into the learning process, ErrP-IRL~\cite{batzianoulis2021} integrated error-related potentials~\cite{millan2008} with IRL. This approach assigns trajectory weights based on users’ cognitive responses, which are then used to iteratively refine a reward function represented as a sum of radial basis functions.

Beyond human feedback, recent works have explored learning reward functions from large-scale, unstructured demonstrations. GraphIRL~\cite{2207.14299} extracted task-specific embeddings from diverse video demonstrations. By modeling object interactions as graphs and performing temporal alignment, GraphIRL learns transferable reward functions without requiring explicit reward design or environment correspondence, enabling cross-domain manipulation capabilities. To further improve policy precision, Naranjo-Campos et al.~\cite{app142311131} proposed to integrate IRL with Proximal Policy Optimization~\cite{schulman2017proximal}. Their method incorporates expert-trajectory-based features and a reverse discount factor to address feature vanishing issues near goal states, thereby improving the robustness of the learned policies. More recently, Visual IRL~\cite{2412.11360} extended the scope of IRL to human-robot collaboration tasks. It employs adversarial IRL to infer reward functions from human demonstration videos and introduces a neuro-symbolic mapping that translates human kinematics into robot joint configurations. This approach not only ensures accurate end-effector placement but also preserves human-like motion dynamics, facilitating natural and effective robot behavior in dexterous manipulation tasks.

\subsubsection{Discussion}


In summary, IRL has demonstrated significant potential for dexterous manipulation tasks. By inferring the underlying reward function from expert demonstrations, IRL enables robots to generalize complex behaviors and adapt to diverse environments without the need for manually designed reward functions. This capability is particularly valuable in dexterous manipulation scenarios where reward specification is challenging or impractical~\cite{ARORA2021103500, s23073762}. Despite these promising developments, state-of-the-art IRL methods still face several limitations. One of the primary challenges lies in accurately estimating reward functions, particularly in environments with high-dimensional action spaces or sparse feedback signals. Furthermore, IRL methods often rely on large amounts of expert demonstration data, which poses practical constraints due to the high cost and time required for data collection~\cite{ARORA2021103500, 10493015}.



\subsection{Generative Adversarial Imitation Learning}
\subsubsection{Description}
Generative Adversarial Imitation Learning (GAIL) extends the Generative Adversarial Network (GAN) framework~\cite{goodfellow2014generative} to the domain of IL. It formulates the imitation process as a two-player adversarial game between a generator and a discriminator. The generator corresponds to a policy \( \pi \) that aims to produce behavior that closely resembles expert demonstrations, while the discriminator \( D(s, a) \) evaluates whether a state-action pair \( (s, a) \) originates from the expert data \( M \) or is generated by \( \pi \).

Specifically, GAIL minimizes the Jensen-Shannon divergence between the state-action distributions of the expert and the generator. The discriminator is trained to maximize the following objective:
\begin{equation}
\arg\min_D -\mathbb{E}_{d^M(s, a)}[\log D(s, a)] - \mathbb{E}_{d^\pi(s, a)}[\log(1 - D(s, a))]  \label{GAIL}
\end{equation}
where \( d^M(s, a) \) and \( d^\pi(s, a) \) denote the state-action distributions of the expert and the generator, respectively. The generator's policy \( \pi \) is optimized using RL with a reward signal derived from the discriminator:
\begin{equation}
r_t = -\log(1 - D(s_t, a_t)).
\end{equation}
Through this adversarial training process, GAIL effectively learns complex behaviors from expert demonstrations without explicitly recovering the reward function.





\subsubsection{Research Progress}
GAIL has been widely adopted in dexterous manipulation. However, its effectiveness heavily relies on the quality and availability of expert demonstrations, which are often labor-intensive to collect and prone to inconsistencies~\cite{tangkaratt2019vild, wang2021learning}. Such discrepancies arise from factors like collector biases~\cite{zuo2021adversarial}, expert errors, noisy data, non-convex solution spaces, and suboptimal strategies~\cite{antotsiou2021adversarial}. Additionally, data scarcity further limits learning efficiency and policy robustness.

To address these challenges, various GAIL extensions have been proposed. HGAIL~\cite{liu2019hindsight} employed hindsight experience replay to synthesize expert-like demonstrations without requiring real expert data. AIL-TAC~\cite{antotsiou2021adversarial} introduced a semi-supervised correction network to refine noisy demonstrations. GAIL has also been used in sim-to-real transfer~\cite{jiang2023mastering}, reducing the dependence on real-world expert data. Nevertheless, GAIL still suffers from mode collapse, where learned policies capture only a narrow range of behaviors, and gradient vanishing issues when the discriminator overpowers the generator. To mitigate these problems, RIDB~\cite{wang2017robust} incorporated variational autoencoders to learn semantic policy embeddings and enable smooth interpolation across behaviors. WAIL~\cite{xiao2019wasserstein} leveraged the Wasserstein GAN framework~\cite{arjovsky2017wasserstein} to improve training stability and reduce mode collapse. DIL-SOGM~\cite{vahabpour2024diverse} further introduced a self-organizing generative model to capture multiple behavioral patterns without requiring encoders. In parallel, several works improve GAIL's robustness under imperfect demonstrations. GA-GAIL~\cite{tsurumine2022goal} employed a second discriminator to identify goal states, enhancing policy learning from suboptimal data. RB-GAIL~\cite{shi2024ranking} integrated ranking mechanisms and multiple discriminators to model diverse behavior modes while leveraging generated experiences.

In addition to addressing the quality and availability of expert demonstrations, recent studies have sought to improve the performance of GAIL-based dexterous manipulation methods in other aspects. For instance, TRAIL~\cite{zolna2021task} introduced constrained discriminator optimization to prevent the discriminator from focusing on spurious, task-irrelevant features such as visual distractors, thereby preserving meaningful reward signals and enhancing task performance. In the context of human imitation, Antotsiou et al.~\cite{antotsiou2018task} combined inverse kinematics and particle swarm optimization with GAIL to mitigate sensor noise and domain discrepancies, enabling robots to autonomously grasp objects in simulation environments. Furthermore, P-GAIL~\cite{tsurumine2019generative} incorporated entropy-maximizing deep P-networks into GAIL to improve policy learning in deformable object manipulation tasks.

\subsubsection{Discussion}
Although several extensions have addressed specific challenges of GAIL in dexterous manipulation, it still inherits the fundamental limitations of adversarial training. In particular, GAIL often suffers from training instability and faces difficulties in scaling to high-dimensional action spaces.


\subsection{Hierarchical Imitation learning}
\subsubsection{Description}
Hierarchical Imitation Learning (HIL) is an IL framework designed to address complex tasks by decomposing them into a hierarchical structure. HIL typically adopts a two-level hierarchy, where the high-level policy is responsible for generating a sequence of sub-tasks or primitives based on the current state and task requirements, and the low-level policy executes sub-tasks to achieve the overall objective. This hierarchical decomposition enables handling long-horizon and complex tasks more effectively by separating decision-making and control. 

Mathematically, the high-level policy \( \pi_h \) selects a primitive \( p_i \) from a predefined set of primitives \( \{p_1, p_2, \dots, p_K\} \):
\[
\pi_h(s_t) = p_i
\]
where $i \in \{1, 2, \dots, K\}$. The corresponding low-level policy \( \pi_{p_i} \) then generates the action to execute the selected primitive:
\[
a_t = \pi_{p_i}(s_t).
\]
The overall objective of HIL is to minimize the cumulative loss function  \( \mathcal{L}(\pi) \), which explicitly reflects the hierarchical structure of the policy by jointly optimizing both the high-level decision-making and the low-level control execution to achieve effective task decomposition and coordination:
\begin{equation}
\mathcal{L}(\pi) = \sum_{t=1}^T \mathbb{E}_{(s_t, a_t) \sim \pi}\left[\ell(s_t, a_t)\right]
\end{equation}
where \( \ell(s_t, a_t) \) represents the immediate loss at time step \( t \).

The parameters of the high-level and low-level policies in HIL are typically determined through three approaches: (1) Learning from demonstrations, which utilizes expert demonstrations to train both levels of policies; (2) Optimization, which applies RL or other optimization methods to refine the policies; and (3) Manual tuning, which manually adjusts policies during the initial stages or for specific task requirements. A key advantage of HIL is its ability to reduce the complexity of direct action-space search by decomposing tasks into hierarchical structures. This decomposition not only improves learning efficiency, particularly in long-horizon tasks, but also enhances generalization and task success rates by enabling optimization at multiple levels.

\subsubsection{Research Progress}
In recent years, HIL has made significant progress in task decomposition and skill generalization. CompILE \cite{Kipf2018CompositionalIL} enhanced generalization in complex environments by decomposing tasks into independent sub-tasks, laying the foundation for subsequent work, particularly for tasks with long temporal dependencies. In dexterous manipulation, ARCH \cite{2409.16451} integrated a low-level library of predefined skills with a high-level IL policy, enabling efficient skill composition and adaptation for complex, high-precision tasks. Meanwhile, Wan et al.~\cite{Wan2023LOTUSCI} emphasized maintaining skill continuity and stability in dynamic environments by decomposing tasks into continuous sub-tasks and enabling adaptive policy transitions under varying conditions.

\color{black}
To enhance scalability, recent research has increasingly explored automatic hierarchy discovery to reduce reliance on manually designed skill structures. Unsupervised skill segmentation methods have shown strong potential for identifying reusable sub-skills and task boundaries directly from raw data. For example, Gehring et al.\cite{gehring2021hierarchical} proposed hierarchical exploration strategies that use unsupervised clustering to learn multi-level reusable sub-policies. More recently, Jansonnie et al.\cite{jansonnie2024unsupervised} presented a skill discovery framework that leverages automatic task generation and asymmetric self-play to produce diverse manipulatory behaviors, which are then embedded into reusable primitives suitable for hierarchical policies. Similarly, contrastive learning approaches \cite{yang2023behavior} discovered skills by maximizing mutual information across trajectory embeddings, resulting in more robust and generalizable skill representations.
\color{black}

From a data-centric perspective, play data has been used to train both high-level and low-level policies. Wang et al.\cite{wang2023mimicplay} proposed MimicPlay, which exploits unstructured human play interactions to learn high-level latent plans and trains a low-level visuomotor controller using only a small number of demonstrations. Lin et al.\cite{10610084} introduced H2RIL, which extracts interaction-aware skill embeddings from task-agnostic play data and aligns them with human videos via temporal contrastive learning, enabling generalization to novel, composable tasks and adaptation to out-of-distribution scenarios.

\subsubsection{Discussion}
In summary, HIL has demonstrated significant advantages in task decomposition, skill generalization, and handling long-horizon tasks. By introducing hierarchical structures and multi-level control strategies, substantial progress has been made across various dexterous manipulation tasks. However, most current approaches still depend on manually designed hierarchies, which limit autonomy in unstructured or dynamic environments. Automatic hierarchy discovery methods, such as unsupervised skill segmentation or contrastive learning-based sub-task discovery, are essential for enabling robots to autonomously identify reusable skills and adapt their hierarchy to new tasks. Furthermore, ensuring robustness and continuity when skill libraries need to be updated or extended, particularly under changing task conditions, remains an open problem.

 \subsection{Continual Imitation Learning}
\subsubsection{Description}
Continual Imitation Learning (CIL) combines continual learning with IL to enable robots to continuously acquire and adapt new skills from expert demonstrations while avoiding the forgetting of previously learned knowledge when adapting to new tasks in dynamic environments. Specifically, in the initial phase, the agent learns fundamental skills from expert demonstrations. In subsequent phases, the agent incrementally accumulates knowledge, adapts to new tasks or environments, and mitigates the risk of forgetting previously acquired skills.

In CIL, the policy \( \pi \) is optimized by minimizing the cumulative imitation loss across all previously encountered tasks. The objective function is defined as:
\begin{equation}
\mathcal{L}(\pi) = -\sum_{i=1}^{t} \lambda(i) \, \mathbb{E}_{(s^{(i)}, a^{(i)}) \sim \rho^{(i)}_{\text{exp}}} \left[ \log \pi\left( a^{(i)} \mid s^{(i)} \right) \right]
\end{equation}
where \( \lambda(i) \) assigns a weight to each of the \( t \) tasks, and \( \rho^{(i)}_{\text{exp}} \) denotes the distribution of expert state-action pairs. The core objective of CIL is to continuously refine the policy $\pi$ using new demonstrations while preserving performance on previously learned tasks. This is particularly challenging, as it requires balancing the acquisition of new skills without compromising the proficiency of previously acquired ones.

\subsubsection{Research Progress}

Various CIL approaches have been developed for dexterous manipulation, each aiming to balance memory retention and adaptation. Early studies~\cite{Liang2024NeverEndingBA} relied on storing large amounts of task data, leading to high storage and computational costs. To address this, task-specific adapter structures introduced lightweight, modular components for seamless task switching~\cite{Liu2023TAILTA}, though their performance declined when task variations are substantial.

\color{black}
To address catastrophic forgetting more systematically, regularization-based methods\cite{zhaostatistical} have gained prominence. For instance, Elastic Weight Consolidation (EWC)\cite{kirkpatrick2017overcoming} penalized large updates to parameters critical for previously learned tasks by incorporating Fisher information into the loss function. Similarly, knowledge distillation-based approaches\cite{li2024continual} preserved prior skills by encouraging new policies to mimic outputs from older policies. Rehearsal-based strategies offered another solution, such as experience replay, where past experiences or generated trajectories are interleaved with new task data to reduce forgetting\cite{rolnick2019experience}. Generative models, such as Deep Generative Replay (DGR), further alleviated storage constraints by synthesizing realistic task trajectories instead of storing raw data~\cite{Gao2021CRILCR}, although ensuring trajectory fidelity remains challenging.

Beyond these, unsupervised skill discovery has been explored to improve adaptability and generalization~\cite{Wan2023LOTUSCI}. It dynamically generated new skills and integrated them into the robot’s repertoire, but had yet to demonstrate robust real-world generalization. Unified policy learning via behavior distillation~\cite{Haldar2023PolyTaskLU} provided an alternative by employing a single shared policy across tasks, eliminating the need for task-specific modules, though at the cost of increased task interference. Lastly, self-supervised learning techniques have shown promise in acquiring transferable skill abstractions without explicit demonstrations ~\cite{Mete2024QueSTSS}.

\subsubsection{Discussion}
In summary, CIL for dexterous manipulation has explored a range of techniques, such as adapter-based methods, regularization approaches, knowledge distillation, rehearsal strategies, and generative models. These approaches collectively aim to mitigate catastrophic forgetting and enhance task-switching efficiency, but significant challenges remain. The effectiveness of regularization and knowledge distillation often depends on carefully balancing new and old task performance, which becomes difficult when tasks differ substantially. Similarly, the stability and performance of rehearsal or generative replay methods are highly sensitive to the quality and diversity of stored or synthesized data. Although modular adapters and regularization techniques have reduced memory demands, computational overhead and scalability continue to pose barriers for real-world deployment.
\color{black}

\begin{table*}[t]
\centering
\renewcommand{\arraystretch}{1.5}
\setlength{\tabcolsep}{5pt}

\arrayrulecolor{black} 
{\color{black} 


\caption{A Comparison of IL Methods in Cost, Feasibility, Efficiency, and Complexity}

\label{tab:imitation_learning_analysis}
\resizebox{\linewidth}{!}{
\begin{tabular}{
    >{\centering\arraybackslash}m{1.5cm}|
    >{\centering\arraybackslash}m{1.8cm}|
    >{\centering\arraybackslash}m{2.2cm}|
    >{\centering\arraybackslash}m{3cm}|
    >{\centering\arraybackslash}m{2.2cm}|
    m{4cm}}
\hline
\textbf{Category} & \textbf{Training Cost} & \textbf{Inference Latency} & \textbf{Sample Efficiency} & \textbf{Computational Complexity} & \textbf{Real-Time Feasibility Notes} \\
\hline

\textbf{BC} &      
Low & 
Very Low (\(<1\) ms) & 
Low–Moderate (requires many demos; improves with online correction) & 
Low & 
Highly suitable for high-DOF real-time control; the main bottleneck is data quality and covariate shift.
\\
\hline

\textbf{IRL} & 
High & 
Low–Moderate (1–5 ms) & 
Moderate–Low & 
High & 
Training is computationally expensive; inference is feasible but may require reward computation if deployed online.
\\
\hline

\textbf{GAIL} & 
High & 
Low–Moderate (2–5 ms) & 
Moderate & 
High & 
Inference is lightweight once trained; training is unsuitable for strict real-time constraints.
\\
\hline

\textbf{HIL} & 
Moderate–High & 
Moderate (2–10 ms) & 
High (can reuse sub-policies) & 
Moderate–High & 
Two-tier policy execution increases latency; optimization is needed for tight loops.
\\
\hline

\textbf{CIL} & 
High & 
Moderate (2–8 ms) & 
Moderate–High & 
Moderate–High & 
Online adaptation may break hard real-time constraints; feasible if adaptation is offloaded.
\\
\hline

\end{tabular}
}
} 
\arrayrulecolor{black} 
\end{table*}

\subsection{Comparative Analysis of Imitation Learning Approaches}
While Table~\ref{tab:imitation_learning_comparison} outlines the general pros and cons of different IL approaches, the challenges inherent to dexterous manipulation amplify these limitations and deserve deeper analysis. Table~\ref{tab:imitation_learning_analysis} summarizes the trade-offs between efficacy and computational complexity across imitation learning paradigms. BC offers a simple, data-efficient solution but is highly vulnerable to distribution shifts, where small deviations can rapidly compound into task failure. IRL addresses this by inferring expert reward functions to improve generalization, though designing meaningful rewards for complex tasks remains difficult and computationally intensive. GAIL combines the strengths of BC and IRL by adversarially matching expert behavior without explicit reward design, yet it often suffers from instability and mode collapse in multi-modal tasks. Structurally, Hierarchical IL builds on these methods by decomposing tasks into modular skill primitives, improving scalability and reusability; however, reliance on hand-crafted hierarchies limits adaptability to novel scenarios. Continual IL addresses the evolving nature of manipulation tasks by supporting incremental skill acquisition, but dynamic contact interactions increase the risk of catastrophic forgetting. In summary, BC is well-suited for short-horizon, data-rich settings, while IRL and GAIL provide more robust generalization when expert intent must be inferred. Hierarchical and continual strategies extend these foundations to support long-term skill composition and adaptation. As detailed in Table~\ref{tab:il_optimization}, the optimization specifics and convergence behavior of imitation learning methods are analyzed.

\begin{table*}[t]
\centering
\renewcommand{\arraystretch}{1.5}
\setlength{\tabcolsep}{5pt}

\arrayrulecolor{black}
{\color{black}

\caption{A Comparison of IL Methods by Optimization and Convergence}
\label{tab:il_optimization}

\begin{tabular}{
    >{\centering\arraybackslash}m{1.4cm}|
    m{3.8cm}|
    m{3.2cm}|
    m{3.5cm}|
    m{4.2cm}}
\hline
\textbf{Method} & \textbf{Key Hyperparameters} & \textbf{Common Optimizers} & \textbf{Regularization Strategies} & \textbf{Convergence Behavior Notes} \\
\hline

\textbf{BC} & Learning rate, batch size, network depth & Adam, RMSprop & L2 weight decay, dropout & Converges quickly; risk of overfitting on small datasets; performance limited by covariate shift. \\
\hline

\textbf{IRL} & Reward learning rate, policy learning rate, discriminator updates & Adam for policy \& reward networks & Entropy regularization, reward clipping & Slower convergence due to reward inference; reward shaping essential for stability. \\
\hline

\textbf{GAIL} & Policy LR, discriminator LR, batch size, rollout length & Adam (policy \& discriminator) & Gradient penalty, spectral norm & Sensitive to discriminator–policy update ratio; prone to oscillation without tuning. \\
\hline

\textbf{HIL} & Subpolicy LR, meta-controller LR, option horizon & Adam (separate for high- \& low-level) & Option dropout, inter-option regularization & Convergence depends on option discovery quality; mitigates long-horizon credit assignment. \\
\hline

\textbf{CIL} & Base LR, regularization strength for knowledge retention, replay buffer size & Adam, SGD with momentum & Elastic Weight Consolidation (EWC), replay-based regularization & Trains indefinitely with periodic adaptation; stability–plasticity trade-off crucial. \\
\hline

\end{tabular}
}
\arrayrulecolor{black}
\end{table*}
\color{black}

In addition to these strengths and weaknesses, ongoing debates highlight critical trade-offs that influence their application in complex dexterous manipulation tasks. (1) A central debate revolves around the balance between simplicity and scalability. BC is simple and data-efficient, making it effective for tasks with lower complexity or shorter horizons, but it struggles with generalization in high-DoF dexterous manipulation tasks. In contrast, IRL and GAIL offer better generalization, with IRL excelling at capturing expert behavior for unseen states but facing scalability issues due to the challenge of reward design. GAIL, on the other hand, provides greater scalability by eliminating the need for explicit reward functions, but it is prone to instability and mode collapse, particularly in high-DoF or multi-modal tasks. (2) Another ongoing discussion centers on the trade-off between end-to-end learning from raw observations and using intermediate representations. End-to-end learning simplifies the pipeline by directly mapping raw inputs to outputs, but it often requires large amounts of data and struggles with generalization, especially in complex tasks. In contrast, using intermediate representations, such as semantic keypoints, can improve efficiency and generalization by focusing on task-relevant abstractions. However, this approach introduces added complexity in terms of feature extraction and domain-specific design, increasing the computational burden. (3) A further debate concerns the trade-off between learning from expert demonstrations versus self-exploration. Learning from expert demonstrations (e.g., via BC or GAIL) can offer highly effective solutions by mimicking human-like actions, but it relies heavily on high-quality expert data, which may be scarce or difficult to obtain. Self-exploration, however, allows the model to learn through trial and error, leading to greater flexibility in task environments. Yet, this approach often suffers from inefficiencies and a longer learning time, as it requires the agent to explore numerous states to achieve a sufficient understanding of task-specific behaviors.

Hybrid strategies hold promise for achieving both robustness and scalability in complex, long-horizon tasks by integrating complementary strengths of different paradigms. A key direction involves integrating hierarchical planning with adversarial imitation by employing a high-level policy that generates symbolic or language-conditioned subgoals, while the low-level policy is optimized through adversarial learning to match expert trajectories under complex contact dynamics. This coupling allows the high-level controller to focus on long-horizon task composition while the low-level controller benefits from the sample efficiency and robustness of adversarial training in modeling fine-grained manipulation primitives. Another approach is embedding continual learning mechanisms into hierarchical architectures, enabling incremental skill acquisition while preserving previously learned behaviors through replay-based regularization or generative memory. Reward shaping derived from IRL can be combined with supervised BC objectives to improve generalization and reduce compounding errors in data-limited settings. Furthermore, hybrid systems can integrate meta-learning for rapid skill adaptation, while lightweight model-based modules, such as MPC for local corrections, complement model-free IL for enhanced stability and precision.

An increasingly important aspect of hybrid IL strategies is the integration of model-based control with imitation learning, which is often underexplored. Model-based components, such as dynamics models or physics priors, can provide structured guidance for low-level policy optimization, reducing reliance on dense supervision or large datasets. For instance, MPC can be used to plan feasible trajectories or correct drift during execution, while the IL policy handles task-level reasoning and adaptation. Conversely, IL can help bootstrap model-based controllers in regions where dynamic models are inaccurate or data is sparse. This synergistic integration allows systems to exploit the strengths of both paradigms: the interpretability, adaptability, and safety of model-based control with the expressiveness and flexibility of data-driven IL. In real-world manipulation, where uncertainties and contact-rich dynamics are prevalent, such hybrid control strategies have shown promise in improving generalization, robustness, and sample efficiency.
\color{black}


\begin{table*}[t]
\centering
\renewcommand{\arraystretch}{1.5}
\setlength{\tabcolsep}{5pt}

\arrayrulecolor{black}
{\color{black}

\caption{Comparative Performance of Three End-Effector Types in Dexterous Manipulation}
\label{tab:end_effector_il}

\resizebox{\linewidth}{!}{

\begin{tabular}{
    >{\centering\arraybackslash}m{2.2cm}|
    >{\centering\arraybackslash}m{0.9cm}|
    m{2.5cm}|
    m{2cm}|
    >{\centering\arraybackslash}m{1.5cm}|
    >{\centering\arraybackslash}m{1.5cm}|
    >{\centering\arraybackslash}m{1cm}|
    m{1.8cm}|
    m{2.3cm}}
\hline
\textbf{End-Effector Type} & \textbf{DoF} & \textbf{Dexterity \& Design Notes} & \textbf{Actuation Principle} & \textbf{Control Precision} & \textbf{Adaptability} & \textbf{Cost} & \textbf{Typical Applications} & \textbf{Suitability for IL} \\
\hline

\textbf{Two-Fingered Gripper} & 1--2 & Simple parallel jaws; limited in-hand re-configuration & Single-motor, linkage or direct-drive & Medium & Low & Low & Pick-and-place, repetitive tasks & Low data demand, easy to train, strong generalization \\
\hline

\textbf{Three-Fingered Robotic Claw} & 6--9 & Three opposed fingers (Fig.\ref{fig:robotic_claws}); hybrid under-actuation yields cylindrical/spherical grasps with modest dexterity & Tendon-driven or linkage + spring coupling & Medium--High & Medium & Medium & General grasping, limited in-hand manipulation & Moderate data demand, balances flexibility and training cost \\
\hline

\textbf{Multi-Fingered Anthropomorphic Hand} & 15--25 & Full five-finger layout (Fig. \ref{fig:multi-fingered_hands}); high DoF enables human-like precision and fingertip control & Remote tendon, direct-drive, or hybrid transmissions & High & High & High & Medical procedures, precision assembly, complex tasks & High data demand, expensive training, limited generalization \\
\hline

\end{tabular}
}
}

\arrayrulecolor{black}

\end{table*}


\section{End-effectors for Dexterous Manipulation}




An end-effector is a component at the tip of a robotic manipulator that directly interacts with the environment to perform tasks. In dexterous manipulation, end-effectors are typically categorized into two-fingered grippers, multi-fingered anthropomorphic hands, and three-fingered robotic claws. This section introduces these three types in order, highlighting their design principles, advantages, and trade-offs.
\color{black}Table \ref{tab:end_effector_il} provides a comparative analysis of the performance characteristics of these three end-effector types in dexterous manipulation.
\color{black}

\subsection{Two-fingered Traditional Gripper}


Two-fingered grippers are widely used for their reliability, simplicity, and ease of control. Typically driven by a single actuator with one DoF, they are cost-effective and suitable for repetitive tasks requiring consistency \cite{yu_dexterous_2022}. For example, \cite{nasiriany2022learningretrievalpriordata} demonstrated a Franka robot with such a gripper performing tasks like setting a breakfast table. Similarly, Kim et al. \cite{kim2020using} used a two-fingered gripper for behavior cloning with gaze prediction, and a tendon-driven variant in \cite{ciocarlie_kinetic_2013} showed the ability to grasp diverse household objects.

Recent works have further extended gripper capabilities through large-scale imitation datasets such as MIME \cite{sharma_multiple_2018}, RH20T \cite{fang_rh20t_2024}, Bridge Data \cite{ebert_bridge_2022}, and Droid \cite{khazatsky_droid_2024}. Dual-arm systems also enhanced manipulation by coordinating two grippers. For instance, \cite{kim_goal-conditioned_2024} achieved banana peeling through dual-action IL. More complex tasks such as shrimp cooking, cloth folding, and dishwashing were demonstrated in Mobile ALOHA \cite{fu2024mobile} and UMI \cite{chi2024universal}.





Despite these advances, two-fingered grippers remain fundamentally limited in dexterous manipulation, which requires within-hand object reconfiguration \cite{ma_dexterity_2011}. Their simple structure and lack of internal DoFs restrict post-grasp adjustments \cite{yu_dexterous_2022}. Furthermore, morphological differences from the human hand hinder learning from demonstrations and prevent replication of human-like in-hand movements \cite{kadalagere_sampath_review_2023}.

\subsection{Multi-fingered Anthropomorphic Hand}


To overcome the dexterity limitations of two-fingered grippers, robotic hands with human-like morphology have been widely developed. These anthropomorphic hands are better suited for interacting with objects and environments designed for humans \cite{vianello_human-humanoid_2021}. They can be typically classified by transmission mechanisms—tendon-driven, linkage-driven, direct-drive, and hybrid systems—which fundamentally affect their performance characteristics \cite{kim_integrated_2021, hu_design_2023}.

\begin{figure}[t]
  \centering
  \includegraphics[width=1.0\linewidth]{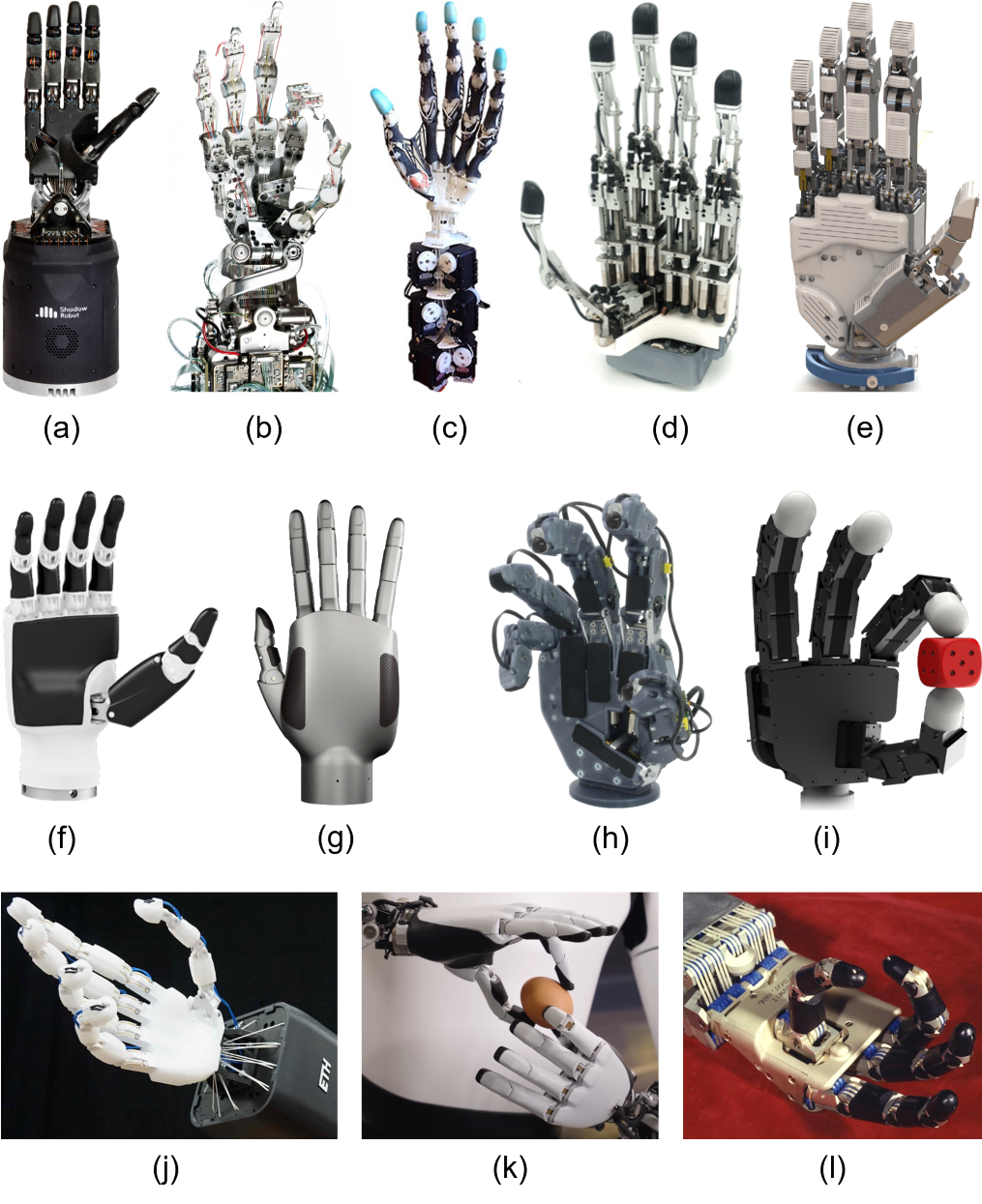}
  \caption{Examples of multi-fingered anthropomorphic hands: (a) Shadow Dexterous Hand\cite{shadow_hand_web2}; (b) Awiwi Hand \cite{Awiwi_Hand_web}; (c) Biomimetic Hand \cite{Biomimetic_Hand_web}; (d) ILDA Hand \cite{kim_integrated_2021}; (e) Hu et al.'s robotic hand \cite{hu_design_2023}; (f) INSPIRE-ROBOTS RH56 Dexterous Hand\cite{inspire_RH56hand_web}; (g) Linker Hand L20 \cite{linkerhand}; (h) PUT-Hand \cite{mankowski_put-handhybrid_2020}; (i) Allegro Hand\cite{Allegro__Hand_web}; (j) Faive Hand \cite{Faive_hand_web}; (k) Tesla Optimus Hand\cite{Tesla_Optimus_web}; (l) Utah/MIT Dexterous Hand \cite{jacobsen_design_1986}.}
  \label{fig:multi-fingered_hands}
\end{figure}

\subsubsection{Tendon-driven Approach}





Tendon-driven hands use cable transmissions to actuate joints, mimicking human tendons. This design allows for compact structure, multiple DoFs, and high dexterity, making it a common choice in anthropomorphic hand development. To accommodate high DoFs, actuators are often remotely located in the forearm.


Representative examples include the Utah/MIT Hand \cite{jacobsen_design_1986} (see Fig. \ref{fig:multi-fingered_hands}(l)), the Shadow Dexterous Hand \cite{shadow_hand_web} (see Fig. \ref{fig:multi-fingered_hands}(a)), and the Awiwi Hand \cite{grebenstein_dlr_2011, grebenstein_hand_2012, grebenstein_antagonistically_2010, grebenstein_antagonism_2008} (see Fig. \ref{fig:multi-fingered_hands}(b)), all of which adopt antagonistic tendon routing for biomimetic motion. The FLLEX Hand \cite{kim_fluid_2019, FLLEX_Hand_v2_web} and Faive Hand \cite{toshimitsu_getting_2023} (see Fig. \ref{fig:multi-fingered_hands}(j)) with rolling contact joints demonstrate robustness and ball-rolling manipulation, respectively. Other typical tendon-driven hands include the Robonaut R2 Hand \cite{bridgwater_robonaut_2012}, Valkyrie Hand \cite{radford_valkyrie_2015, guo_design_2014}, UB Hand \cite{lotti_development_2005, palli_development_2012, melchiorri_development_2013}, DEXMART Hand \cite{palli_dexmart_2014, siciliano_advanced_2012}, iCub Hand \cite{sureshbabu_new_2015}, and Biomimetic Hand \cite{zhe_xu_design_2016} (see Fig. \ref{fig:multi-fingered_hands}(c)).

While remote actuation reduces hand weight, it introduces friction and tendon wear due to long transmission paths. To address this, some designs embed all actuators within the palm. Examples include the DEXHand \cite{chalon_dexhand_2011}, SpaceHand \cite{chalon_spacehand_2015}, CEA Hand \cite{martin_design_2014}, and OLYMPIC Hand \cite{liow_olympic_2020}, which prioritize modularity and compact integration. Commercial designs like DexHand 021 \cite{DexRobot_DexHand021_web}, Tesla Optimus Hand \cite{leddy_underactuated_2024} (see Fig. \ref{fig:multi-fingered_hands}(k)) and PUDU DH11 Hand \cite{PUDU_DH11_hand_web} also follow this approach.

Despite their advantages in dexterity and anthropomorphism, tendon-driven hands face challenges such as friction loss \cite{reinecke_guiding_2014, uchiyama_method_1995}, end termination \cite{grebenstein_antagonistically_2010, grebenstein_approaching_2014, gerez_compact_2018}, tendon creep and wear \cite{palli_modeling_2012, friedl_frcef_2015, goswami_dlr_2017}, which impact durability and reliability. As a result, most remain within research settings, with limited deployment in real-world industrial applications.

\subsubsection{Linkage-driven Approach}



Linkage-driven hands use rigid mechanical linkages to control joint motion, offering high precision, repeatability, and robustness. Compared to tendon-driven designs, they generally provide fewer DoFs but benefit from simpler, more reliable actuation. As a result, most commercial prosthetic and robotic hands adopt this mechanism. Due to space constraints and the demand for compactness, most linkage-driven fingers are actuated by a single motor and fall into two main categories: one-DoF coupled and multi-DoF underactuated types \cite{kashef_robotic_2020}.

In the one-DoF type, joints are mechanically coupled, so preshaping remains fixed during flexion. Typical designs include the S-finger with inverse four-bar coupling \cite{imbinto_s-finger_2018} and the humanoid hand by Liu et al. using two four-bar linkages per finger \cite{liu_development_2016}. Similar configurations appear in hands like the INSPIRE-ROBOTS RH56 \cite{inspire_RH56hand_web} (see Fig. \ref{fig:multi-fingered_hands}(f)), Bebionic Hand \cite{bebionic_Hand_web, medynski_bebionic_2011}, BrainRobotics Hand \cite{brainrobotics_Hand_web} and OYMotion OHand \cite{OYMotion_OHand_web}.



In contrast, underactuated fingers can adapt to contact forces, enhancing grasp adaptability. Examples include the Southampton Hand with a Whiffle tree mechanism \cite{kyberd_design_2001}, the LISA Hand using linkage-based self-adaptation \cite{jin_lisa_2012}; MORA HAP-2 Hand \cite{gopura_prosthetic_2017}, AR Hand III \cite{yang_anthropomorphic_2009}, and Cheng et al.'s prosthetic hand \cite{cheng_design_2017} with multi-bar or four-bar adaptive linkages.


While most designs use one motor per finger, a few incorporate multiple actuators for higher dexterity. The ILDA Hand \cite{kim_integrated_2021} (see Fig. \ref{fig:multi-fingered_hands}(d)) employs three motors per finger with combined PSS/PSU chains and four-bar linkages, achieving workspace and fingertip force comparable to human hands. Similar high-DoF linkage designs appear in the Linker Hand L20~\cite{linkerhand}(see Fig. \ref{fig:multi-fingered_hands}(g)), AIDIN ROBOTICS Hand \cite{aidinRobotics_Hand_web} and the RY-H1 Hand 
\cite{Ruiyan_RYH1_Hand_web}.

\subsubsection{Direct-driven Approach}

Direct-drive hands eliminate intermediate transmission by connecting actuators directly to joints. This simplifies the mechanical structure while still allowing for high actuatable DoFs, similar to tendon-driven designs.



Representative examples include the OCU-Hand \cite{mahmoud_dexterous_2010} with 19 DoFs, where most joints are individually driven by embedded DC motors, and the TWENDY-ONE hand \cite{iwata_design_2009}, which achieves 13 DoFs via joint-level motor placement. The KITECH-Hand \cite{lee_kitech-hand_2017}, Allegro Hand \cite{Allegro__Hand_web} (see Fig. \ref{fig:multi-fingered_hands}(i)) and LEAP Hand \cite{shaw_leap_2023} adopt modular finger designs, integrating motors directly into the phalanges. The LEAP Hand also introduces a novel universal abduction-adduction motor configuration for enhanced MCP joint flexibility.

While direct drive offers high control precision and responsiveness, it introduces potential drawbacks such as increased mass, rotational inertia, and finger bulkiness, which may hinder agility in fine manipulation tasks. These limitations partly explain why most direct-drive hands adopt a four-finger configuration.

\subsubsection{Hybrid-transmission Approach}



In addition to the transmission types introduced above, many anthropomorphic hands adopt hybrid schemes to integrate the advantages of different approaches.


For example, the DLR/HIT Hand II \cite{liu_multisensory_2008} and NAIST Hand \cite{ueda_development_2005} use modular fingers with a combination of motors, belts, gears, tendons, or linkage systems. The MCR-Hand series \cite{yang_affordable_2021, yang_low-cost_2021} utilizes a linkage-tendon mixed transmission system to achieve compactness and high functionality. Adab Mora Hand \cite{abayasiri_under-actuated_2020}, LEAP Hand V2 (DLA Hand) \cite{shaw_leap_2024}, and Hu et al.'s hand \cite{hu_design_2023} (see Fig. \ref{fig:multi-fingered_hands}(e)) also integrate multiple transmission elements within fingers to enhance overall performance and adaptability. 


Other hybrid designs explicitly differentiate mechanisms across fingers to match specific functional needs. The PUT-Hand \cite{mankowski_put-handhybrid_2020} (see Fig. \ref{fig:multi-fingered_hands}(h)), for instance, combines a direct-drive thumb, linkage-driven index/middle fingers, and tendon-driven ring/little fingers. Similarly, the MPL v2.0 Hand \cite{johannes_overview_2011}, Tact Hand \cite{slade_tact_2015}, and Six-DoF Open Source Hand \cite{krausz_design_2016} adopt direct or geared actuation for the thumb while using tendons, linkages or timing belts for other fingers. Hands developed by Owen et al. \cite{owen_development_2018}, Ryu et al. \cite{ryu_development_2020}, and Ke et al. \cite{ke_new_2021} follow a similar approach, implementing thumb-specific hybrid strategies to enhance opposability and dexterity.

\begin{table*}[t]
\centering
\renewcommand{\arraystretch}{1.5}
\setlength{\tabcolsep}{5pt}

\arrayrulecolor{black}
{\color{black}

\caption{Key Characteristics of Representative Robotic Hands}
\label{tab:robotic_hands}

\resizebox{\linewidth}{!}{

\begin{tabular}{
    >{\centering\arraybackslash}m{3.0cm}|
    >{\centering\arraybackslash}m{2.2cm}|
    m{3.2cm}|
    m{3.5cm}|
    m{4.2cm}}
\hline
\textbf{Robotic Hand} & \textbf{DoF (actuated / total)} & \textbf{Actuation Type} & \textbf{Key Feature} & \textbf{Typical Applications} \\
\hline

\textbf{Shadow Dexterous Hand\cite{shadow_hand_web2}} & 20 / 24 & Remote tendon-driven (20 Smart Motors in forearm) & Fully biomimetic; 129 sensors; $\pm$1° joint precision & High-dexterity research, tele-operation, AI/ML benchmarking \\
\hline

\textbf{Allegro Hand\cite{Allegro__Hand_web}} & 16 / 16 & Direct-driven (Maxon DC motors in each phalanx) & Modular finger units; ROS-native; low backlash & Academic RL/IL experiments, sim-to-real transfer \\
\hline

\textbf{Linker Hand (L20)\cite{linkerhand}} & 16 / 21 & Linkage-driven (4 motors per finger) & ROS-native; piezoresistive sensors; optional visual-tactile perception & High-dexterity research, Industrial applications \\
\hline

\textbf{BarrettHand (BH8-280)\cite{BarrettHand_web}} & 4 / 7 & Linkage-driven (1 motor per finger + spread) & Under-actuated adaptive fingers; rugged mechanics & Industrial grasping, pick-and-place, educational labs \\
\hline

\textbf{LEAP Hand\cite{shaw_leap_2023}} & 16 / 16 & Direct-driven (compact joint motors) & Low-cost, open-source, anthropomorphic yet lightweight & Learning-focused projects, mobile manipulation \\
\hline

\textbf{i-HY Hand (3-finger)\cite{odhner_compliant_2014}} & 9 / 9 & Tendon-driven (elastic tendons + capstan) & Compliant, under-actuated; high-impact robustness & Robust in-hand manipulation, field robotics \\
\hline

\end{tabular}
}
}
\arrayrulecolor{black}

\end{table*}


\subsection{Three-fingered Robotic Claw: A Trade-off Solution}



The diversity of anthropomorphic hand designs largely stems from a fundamental trade-off between mechanical simplicity and dexterous capability \cite{bicchi_hands_2000}. While the human hand has over 20 DoFs \cite{pena_pitarch_virtual_2008, yang_development_2008, samadani_multi-constrained_2012, lenarcic_robot_2013, prabaharan_nadine_2017, zhou_bcl-13_2018, zarzoura_investigation_2019}, replicating this complexity mechanically remains impractical. Higher dexterity often increases structural and control complexity, cost, and susceptibility to failure \cite{odhner_compliant_2014, demers_kinematic_2011, mnyusiwalla_new_2016}, limiting the feasibility of high-DoF hands in real-world applications \cite{feix_metric_2013, wang_dorahand_2022}.

To mitigate these issues, several simplification strategies are adopted: underactuation with elastic components \cite{demers_kinematic_2011, mnyusiwalla_new_2016, shadow_hand_web2, Tesla_Optimus_web, kyberd_design_2001, jin_lisa_2012}, reducing non-essential DoFs or phalanges \cite{Tesla_Optimus_web, inspire_RH56hand_web, bebionic_Hand_web, OYMotion_OHand_web}, or even omitting a finger entirely \cite{jacobsen_design_1986, radford_valkyrie_2015, chalon_dexhand_2011, iwata_design_2009, shaw_leap_2023, Allegro__Hand_web}. These approaches highlight the challenge of maximizing functionality within practical constraints.

\begin{figure}[ht]
  \centering
  \includegraphics[width=1.0\linewidth]{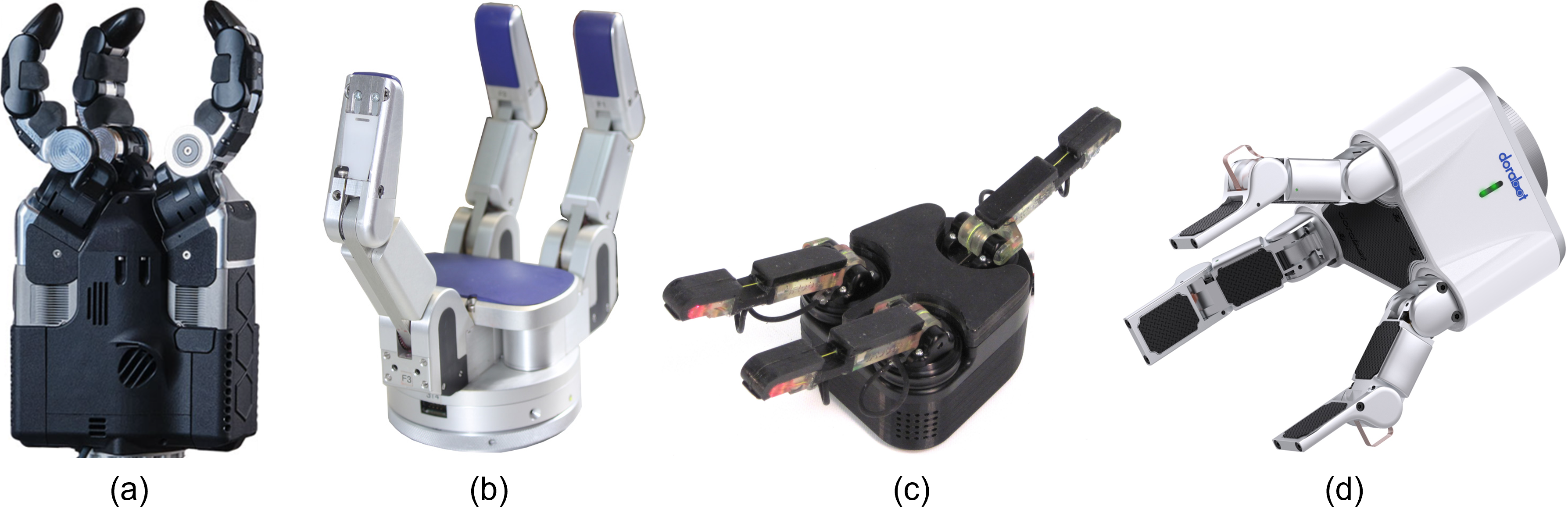}
  \caption{Examples of three-fingered robotic claws: (a) DEX-EE \cite{Shadow_DEXEE_web}; (b) BarrettHand \cite{BarrettHand_web}; (c) i-HY Hand \cite{odhner_compliant_2014}; (d) DoraHand \cite{DoraHand_web}.}
  \label{fig:robotic_claws}
\end{figure}



As a compromise between the minimalist two-fingered gripper and complex multi-fingered anthropomorphic hands, the three-fingered robotic claw offers a functional middle ground. Though not anatomically human-like, three fingers are sufficient for executing common grasp types such as cylindrical and spherical power grasps \cite{odhner_compliant_2014}, and can support a subset of in-hand manipulation tasks.


Numerous three-fingered claws have demonstrated impressive capabilities. For example, Shadow’s DEX-EE \cite{Shadow_DEXEE_web} (see Fig. \ref{fig:robotic_claws}(a)) and the TRX Hand \cite{hu_dexterous_2023} exhibited high robustness and dexterity. The BarrettHand \cite{townsend_barretthand_2000} (see Fig. \ref{fig:robotic_claws}(b)) achieved adaptive grasping through underactuation. Tendon-driven designs like the i-HY Hand \cite{odhner_compliant_2014} (see Fig. \ref{fig:robotic_claws}(c)) and Model O \cite{ma_yale_2017} enabled actions such as pivoting and precision transitions. Systems such as DClaw \cite{zhu_dexterous_2019} and TriFinger \cite{wuthrich_trifinger_2021} were capable of performing fine manipulation tasks via RL. Other novel architectures include linkage-based \cite{li_linkage-driven_2024}, motor-multiplexed \cite{xu_muxhand_2024}, and link-belt-integrated claws \cite{kim_blt_2020}, each offering different characteristics.

Three-fingered claws such as the DoraHand \cite{wang_dorahand_2022} (see Fig. \ref{fig:robotic_claws}(d)), SARAH \cite{laliberte_underactuation_2001}, D’Manus \cite{bhirangi_all_2024}, and Kinova Jaco's claw \cite{Kinova_arm_web} further demonstrated the practicality and versatility of this design choice in both research and assistive applications.\color{black} Based on the foregoing analysis, the key characteristics of representative robotic hands across the three end-effector categories are summarized in Table \ref{tab:robotic_hands}.
\color{black}


\color{black}
\subsection{Tactile Sensors on Dexterous End-effectors}

Tactile sensing is important for enhancing the perception and control capabilities of robotic end-effectors, particularly in dexterous manipulation tasks that require fine contact interaction, compliance, and force regulation. Although previous end-effectors such as two-finger grippers and parallel jaw mechanisms often rely solely on vision or position feedback, modern multi-fingered dexterous hands \cite{ueno2024multi, lin2025pp} are increasingly incorporating high-resolution tactile sensors to achieve more stable and versatile manipulation in unstructured or dynamic environments. 

In general, tactile sensors provide direct measurements of contact force, pressure distribution, slip, surface texture, and sometimes even temperature or material properties. They are typically mounted on the fingertips, phalanges, or palm surfaces of robotic hands, allowing feedback-rich control policies such as force modulation, slip prevention, and adaptive grasping. In recent years, tactile sensors have become essential for learning-based approaches, where tactile data are used for contact state estimation, affordance recognition, and policy refinement~\cite{lee2024dextouch,huang2025dih,heng2025vitacformer,zhang2025kinedex}. 

Tactile sensors can be broadly categorized into several types based on sensing principles:

\begin{itemize}
    \item Resistive and Capacitive Tactile Sensors: These are thin, flexible, and cost-effective, and detect pressure changes through variations in electrical resistance or capacitance. They are widely used in fingertip arrays and robotic skin~\cite{hailiang2024flexible,liu2024material, chen2025capacitive}.
    \item Piezoelectric Sensors: These generate voltage in response to applied stress and are suitable for dynamic contact detection, such as impact and slip sensing~\cite{wang2025piezotac, tang2025recent}.
    \item Optical and Vision-Based Tactile Sensors: Examples such as GelSight~\cite{yuan2017gelsight}, TacTip~\cite{ward2018tactip}, and GelStereo tip~\cite{zhang2025gelstereo} use internal cameras and deformable gel surfaces to reconstruct high-resolution 3D contact geometry. These sensors offer rich, interpretable data and are increasingly used in learning-based manipulation~\cite{sun2025tactile}.
    \item Magnetic and Hall Effect Sensors: Often embedded within soft materials or joints, these detect local deformations or forces via magnetic field changes~\cite{yan2021soft, yan2022tactile, park2023magtac,ding2024soft}.
    \item Bio-inspired Sensors: Inspired by human skin or mechanoreceptors, these sensors aim to capture multi-modal tactile features (e.g., pressure + vibration + shear) in compact designs. Examples include BioTac~\cite{wettels2014multimodal}, NeuTouch~\cite{taunyazov2020event}, and GTac~\cite{lu2022gtac}.
\end{itemize}

The integration of tactile sensing into robotic hands offers significant advantages. First, it enables robust manipulation under conditions of visual occlusion or low illumination by relying solely on contact feedback. Second, tactile sensors facilitate adaptive grasping strategies such as regrasping, force-limited lifting, and slip compensation. Third, tactile data contribute to improved grasp quality estimation, object property inference, and planning for in-hand manipulation. However, tactile sensing also faces several challenges. Calibration and drift over time, especially in soft or flexible sensors, can reduce measurement fidelity. High-resolution sensors (e.g. GelSight \cite{yuan2017gelsight}) are relatively bulky and have limited durability under repeated contact. Furthermore, real-time integration of tactile data into control pipelines demands efficient signal processing, accurate contact modeling, and often machine learning-based interpretation.

In the context of IL for dexterous manipulation, tactile sensing provides a rich stream of contact information that complements visual and proprioceptive cues. Demonstrations collected from expert teleoperation or kinesthetic teaching can include synchronized tactile signals, allowing the learning agent to capture subtle contact events, such as incipient slip, rolling, or micro-adjustments of finger force, that are often invisible to vision alone. By incorporating tactile feedback into the policy representation, IL systems can better generalize to novel objects and adapt to variations in shape, compliance, or surface texture. Recent work ~\cite{huang2025dih, ablett2024multimodal,guzey2024see, murooka2025tact} showed that tactile-informed IL can substantially improve grasp stability, reduce failure rates under visual occlusion, and enable fine in-hand reorientation without explicit object models.

Overall, tactile sensors are becoming an indispensable component of dexterous robotic end-effectors, bridging the gap between perception and control. Future research is expected to focus on developing scalable, durable, and multimodal tactile arrays, as well as fusing tactile signals with vision and proprioception to support generalizable manipulation policies in open-world environments.

\subsection{Impact of End-effector Design on Imitation Learning Performance}

While prior research has primarily focused on algorithmic innovations, the influence of end-effector morphology, actuation, and sensor configuration on IL performance has received comparatively limited attention. These factors can significantly affect data efficiency, policy generalization, and task success rate through the following mechanisms.

\subsubsection{Morphology} 
\textit{Redundant Degrees of Freedom and Ergonomic Alignment.}
High-DoF hands (e.g., Shadow Hand, 20+ DoF) can replicate fine human manipulations such as fingertip control, but they increase the dimensionality of the action space, leading to severe sparsity issues in IL. The RH20T dataset~\cite{fang_rh20t_2024} showed that the demonstration data requirement for high-DoF hands grows exponentially. In contrast, three-fingered grippers (e.g., DoraHand~\cite{wang_dorahand_2022}) traded off some dexterity for reduced policy learning complexity, achieving better zero-shot generalization in cross-domain tasks such as those in BridgeData V2~\cite{walke2023bridgedata}.
Anthropomorphic hands (e.g., ILDA Hand~\cite{kim_integrated_2021}) with geometric consistency to the human demonstrator’s workspace reduced cross-domain mapping errors. Antotsiou et al.~\cite{antotsiou2018task} highlighted that differences in morphology, degrees of freedom, and joint constraints between the human hand and the robot hand can amplify retargeting errors and consequently raise the likelihood of task failure.

\subsubsection{Actuation}
\textit{Trade-Offs Between Transmission Dynamics and Controllability.}
Hands such as the Shadow Hand use tendon-driven actuation, where compliance introduces nonlinear friction and hysteresis, increasing noise in the action–state mapping of demonstration data. 
Tendon-driven systems introduce nonlinearities such as friction and hysteresis in the transmission chain, which degrades control fidelity in dexterous manipulators~\cite{grebenstein_approaching_2014}.
Rigid linkage actuation (e.g. BarrettHand \cite{BarrettHand_web}) reduces dynamic errors, but underactuated designs (e.g. adaptive grasping) limit the reproduction of fine finger motions.
Direct-drive hands (e.g. LEAP Hand \cite{shaw_leap_2023}) achieve millimeter-level control precision through joint-level motors but suffer from increased inertia due to motor mass, requiring IL policies to learn dynamic feedforward compensation. 

\subsubsection{Sensor Configuration}
\textit{Complementarity in Multimodal Perception.}
High-resolution tactile sensing (e.g., GelSight ~\cite{yuan2017gelsight}) can capture micro-force adjustments and slip signals in human demonstrations, significantly improving IL’s understanding of contact dynamics. Huang et al.~\cite{huang2025dih} reported that removing tactile input causes task success rates to plummet to near-random levels, underscoring the critical importance of tactile information for successful multi-object in-hand manipulation.
Differences in sampling frequency (e.g., vision at 30 Hz vs. tactile sensing at 1 kHz) introduce synchronization challenges. Lin et al. [285] addressed this via a cross-modal alignment network, reducing motion consistency error by 25\% in bimanual coordination tasks.
NeuralFeels~\cite{suresh2024neuralfeels} aligned vision with tactile streams via global timestamps and fuses both modalities in a joint neural implicit field, cutting trajectory-consistency error by 25\% for precise in-hand manipulation.

Future end-effector designs should follow a “task–morphology–algorithm” co-optimization paradigm. High-dexterity tasks (e.g., surgical suturing) favor fully actuated hands with dense tactile arrays (e.g., Shadow Dexterous Hand~\cite{shadow_hand_web2}) and employ hierarchical IL frameworks~\cite{kim2025srt} to mitigate the curse of dimensionality. Low-cost generalization tasks (e.g., household tidying) use three-fingered grippers with vision-dominant sensing, combined with domain randomization~\cite{tobin2017domain} to improve cross-object robustness. Real-time collaboration tasks (e.g., human–robot co-manipulation) balance actuation latency and control precision. For instance, RAPID Hand~\cite{wan2025rapid} achieved $<7$ ms response delay through a lightweight design.
\color{black}

\begin{table*}[t]
\centering
\renewcommand{\arraystretch}{1.5}
\setlength{\tabcolsep}{5pt}

\arrayrulecolor{black} 

{\color{black} 

\caption{Comparison of Teleoperation-based Data Collection Approaches}

\label{tab:teleoperation_approaches}


\begin{tabular}{>{\centering\arraybackslash}m{3cm}|m{4.2cm}|m{4cm}|m{5cm}}
\hline
\raisebox{0.6ex}{\textbf{Approach}} \rule{0pt}{4ex} & 
\raisebox{0.6ex}{\textbf{Key Characteristics}} \rule{0pt}{4ex} & 
\raisebox{0.6ex}{\textbf{Advantages}} \rule{0pt}{4ex} & 
\raisebox{0.6ex}{\textbf{Limitations}} \rule{0pt}{4ex} \\
\hline

\textbf{Vision-Based Systems} \cite{8794277,handa2020dexpilot,sivakumar2022robotic,arunachalam2023dexterous,li2022dexterous,qin2022one,qin2023anyteleop,yang2024ace,wang2023mimicplay,mandlekar2018roboturk} & 
\parbox[b]{\hsize}{\vspace{1.5pt}
\begin{itemize}[nosep,leftmargin=3mm]
    \item Non-contact image-based motion capture
    \item Single or multi-camera configurations
    \item RGB or RGB-D inputs, frequently accompanied by hand pose estimation
\end{itemize}
} & 

\parbox[b]{\hsize}{\vspace{1.5pt}
\begin{itemize}[nosep,leftmargin=3mm]
    \item Low-cost hardware (especially single camera)
    \item Easy to deploy and user-friendly
    \item No wearables required
\end{itemize}
} & 

\parbox[b]{\hsize}{\vspace{1.5pt}
\begin{itemize}[nosep,leftmargin=3mm]
    \item Susceptible to occlusion and lighting conditions
    \item Limited 3D precision and depth estimation
    \item Latency due to vision processing
\end{itemize}
} \\
\hline

\textbf{Mocap Gloves} \cite{caeiro2021systematic,wang2024dex,mosbach2022accelerating} & 
\parbox[b]{\hsize}{\vspace{1.5pt}
\begin{itemize}[nosep,leftmargin=3mm]
    \item Wearable gloves with IMUs, flex sensors, or magnetic sensors
    \item Real-time hand joint tracking
    \item Often used with external cameras or LiDAR
\end{itemize}
} & 
\parbox[b]{\hsize}{\vspace{1.5pt}
\begin{itemize}[nosep,leftmargin=3mm]
    \item High joint-level accuracy (millimeter-level)
    \item Fast response time 
    \item Robust under varied visual conditions
\end{itemize}
} & 
\parbox[b]{\hsize}{\vspace{1.5pt}
\begin{itemize}[nosep,leftmargin=3mm]
    \item High equipment cost
    \item Requires calibration and setup
    \item Potential discomfort for extended use
\end{itemize}
} \\
\hline

\textbf{VR/AR Controllers} \cite{zhang2018deep,arunachalam2023holo,radosavovic2023real,mosbach2022accelerating,ding2024bunny,cheng2024opentelevision,lin2024learning} & 
\parbox[b]{\hsize}{\vspace{1.5pt}
\begin{itemize}[nosep,leftmargin=3mm]
    \item Head-mounted display with spatial tracking
    \item Hand-held controllers with 6-DoF input
    \item Immersive virtual environments
\end{itemize}
} & 
\parbox[b]{\hsize}{\vspace{1.5pt}
\begin{itemize}[nosep,leftmargin=3mm]
    \item Intuitive and interactive control
    \item Multimodal feedback (e.g., visual and occasional haptic)
    \item Consumer-grade accessibility
\end{itemize}
} & 
\parbox[b]{\hsize}{\vspace{1.5pt}
\begin{itemize}[nosep,leftmargin=3mm]
    \item Dependence on virtual environments
    \item Challenging deployment and non-negligible control latency
    \item Limited haptic realism and force feedback
\end{itemize}
} \\
\hline

\textbf{Exoskeleton and Bilateral Systems} \cite{falck2019vito,falck2018human,fang2024airexo,kim2023training,zhao2023learning,fu2024mobile,wu2024gello,chi2024universal} & 
\parbox[b]{\hsize}{\vspace{1.5pt}
\begin{itemize}[nosep,leftmargin=3mm]
    \item Direct joint-space mapping
    \item Includes bilateral force and torque feedback
    \item Leader-follower robot configurations
\end{itemize}
} & 
\parbox[b]{\hsize}{\vspace{1.5pt}
\begin{itemize}[nosep,leftmargin=3mm]
    \item High-fidelity control suitable for precision tasks
    \item Enables tactile interaction and real-time feedback
\end{itemize}
} & 
\parbox[b]{\hsize}{\vspace{1.5pt}
\begin{itemize}[nosep,leftmargin=3mm]
    \item Complex and bulky setups
    \item Expensive hardware and maintenance
    \item Restricted operator mobility
\end{itemize}
} \\
\hline
\end{tabular}

} 
\arrayrulecolor{black} 

\vspace{0.1in}
\end{table*}

\section{Dexterous Manipulation with Teleoperation and Video Demonstration}

\color{black}
This section discusses two main approaches to dexterous manipulation: acquiring demonstration data through teleoperation and learning directly from demonstration videos.
Teleoperation facilitates direct human control over robotic systems, leveraging human expertise for learning complex manipulation tasks, while video-based learning harnesses rich visual data to enable autonomous skill acquisition from natural human demonstrations. Together, these approaches address challenges in teaching robots fine-grained manipulation behaviors across diverse scenarios. The section further discusses datasets and benchmarks in supporting imitation learning frameworks that underpin these methodologies.
\color{black}

Teleoperation systems provide a robust interface for human-robot collaboration, benefiting from directly making robot behaviors comply with human-level intelligence, which refers to \textit{human-in-the-loop}. This approach is highly intuitive since humans' extensive knowledge and experience empower them to make informed judgments on diverse tasks across complex scenes and to promptly adjust strategies in response to feedback. Due to this usability, teleoperation is widely applied in various fields.
Additionally, by collecting data on the robot's states and corresponding actions during teleoperation, datasets can be constructed to perform end-to-end IL.

\color{black}
Learning dexterous manipulation skills from demonstration videos offers a promising alternative to teleoperation by leveraging abundant visual data to teach robots complex behaviors. Unlike direct control, video-based learning enables robots to observe human experts performing tasks in diverse and unstructured environments, capturing rich contextual and temporal information. Advances in computer vision and representation learning facilitate extracting meaningful features from raw videos, which can be mapped to robot actions through imitation learning frameworks. This approach not only reduces the reliance on specialized teleoperation hardware but also broadens the scalability of training data, ultimately enabling robots to acquire fine-grained manipulation capabilities from natural human demonstrations.
\color{black}

\subsection{Teleoperation Systems for Dexterous Manipulation}
\begin{figure*}[t]
  \centering
  \includegraphics[width=17cm, height=4cm]{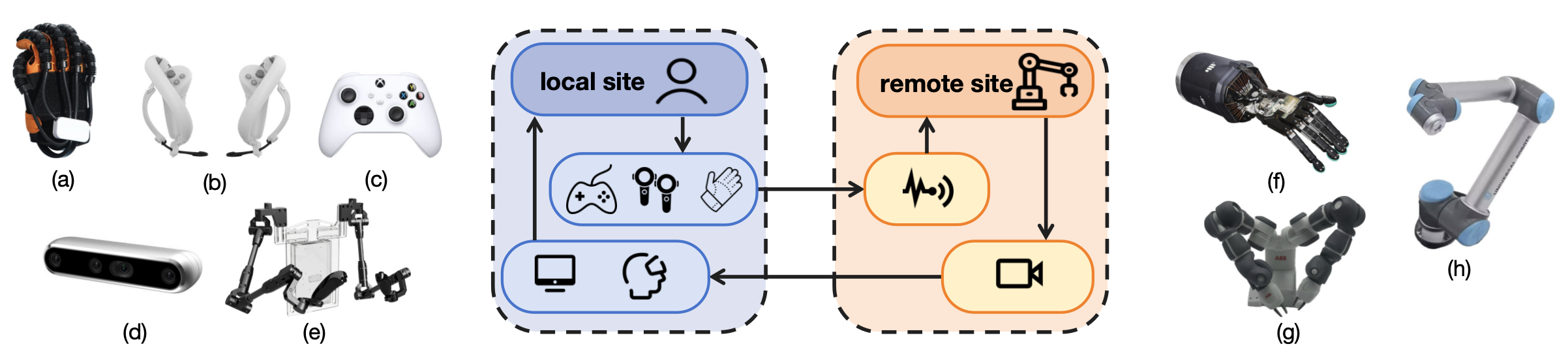}
  \caption{Teleoperation frameworks and commonly used devices: (a) mocap gloves, (b) VR controllers, (c) joystick, (d) RGB-D camera, (e) exoskeleton, (f) dexterous hand, (g) dual-arm robot, (h) single-arm robot.}
  \label{fig:teleoperationframework}
\end{figure*}

A typical teleoperation system consists of two main components: the local site and the remote site, as demonstrated in Fig.~\ref{fig:teleoperationframework}. The local site includes a human operator and a suite of interactive I/O devices. The output devices provide real-time status about the remote robot and its surrounding environment, while the input devices allow the operator to issue commands in diverse forms, thereby controlling the remote robot's actions. The remote site primarily contains the robot itself, which is equipped with various sensors to gather perceptions of its state and the surrounding environment. Upon receiving teleoperation commands from humans, the robots can perform the corresponding actions and complete tasks. 

To accurately convey human operators' intentions to robotic systems, previous works have employed a wide range of human-robot interaction devices. 
Human operators with work experience can easily identify the current state of a robot through the image; however, accurately translating human instructions to robot actions remains a challenge. Some traditional controllers act on this:
1) joysticks \cite{1041657}
2) haptic devices \cite{toedtheide2023force};
However, manipulation tasks often involve delicate movements and complex interactions, such as grasping, moving, and positioning small or irregularly shaped objects. These tasks necessitate devices that can offer dexterous interfaces to ensure the safety and efficacy of the robot's actions. Precision and real-time feedback are crucial.
Commonly used devices include:
1) cameras \cite{handa2020dexpilot,song2020grasping,sivakumar2022robotic,
    qin2023anyteleop,yang2024ace,8794277,li2022dexterous,arunachalam2023dexterous,qin2022one};
2) mocap gloves \cite{6696638,10.1145/3197517.3201399,caeiro2021systematic,kumar2015mujoco,li2020mobile,taheri2020grab};
3) VR/AR controllers \cite{hedayati2018improving,seo2023deep,zhang2018deep,duan2023ar2d2trainingrobotrobot,jang2022bc,arunachalam2023dexterous,arunachalam2023holo,rajeswaran2018learning,ding2024bunny,8003431,8594043,9340822};
4) exoskeletons and bilateral systems \cite{falck2019vito,fang2024airexo,kim2023training,zhao2023learning,fu2024mobile,wu2024gello}.
\color{black}
A comparative summary of representative teleoperation approaches is presented in Table \ref{tab:teleoperation_approaches}, highlighting their key characteristics, advantages, and limitations to facilitate effective human-robot interaction.
\color{black}

\subsubsection{Vision-based Teleoperation Systems}
Recently, advancements in computer vision have led to the development of vision-based teleoperation systems. However, their accuracy in capturing hand movements is often compromised by factors such as occlusion, lighting, resolution, background, and inaccurate 3D estimation issues.
Several methods have been proposed for robust hand pose estimation and reliable mapping to the robot end-effector.
Li et al. \cite{8794277} developed a vision-based teleoperation system by training TeachNet on pairs of images of human hands and simulated robots to form mappings between a human hand and a robotic Shadow Hand in the latent space.
Dexpilot \cite{handa2020dexpilot} utilized a calibrated multi-camera system to estimate hand poses to teleoperate an Allegro Hand. Riemannian Motion Policies (RMPs) are employed to compute the Cartesian pose of the hand, facilitating hand-arm motion control.
Subsequent approaches, such as Robotic Telekinesis \cite{sivakumar2022robotic} and DIME~\cite{arunachalam2023dexterous}, simplified requirements to a single RGB camera, thus reducing the need for calibration. This is achieved through a general mapping method between humans and robots that have different kinematic structures. Additionally, Robotic telekinesis \cite{sivakumar2022robotic} adjusted the position and orientation of the end-effector relative to its base using the relative position and direction of the human wrist to the torso, enabling the teleoperation of both arm and hand. However, these methods still suffer from occlusion issues due to the single fixed camera setting.
To solve this problem, Transteleop~\cite{li2022dexterous} introduced a system that utilizes real-time active vision with a depth camera mounted on the end-effector of the remote UR5 robot arm. During teleoperation, this robot arm can reposition the camera to enhance its field of view and improve hand pose estimation accuracy.

The morphology discrepancy between the human hand and the robot hand might impede the operator from intuitively controlling the robot. 
To address this, Qin et al.~\cite{qin2022one}  developed a user-friendly interface by constructing a customized robotic hand modeled after the specific shape of a human hand. Demonstrations performed with this customized robot hand can be directly transferred to any dexterous robot hand.
AnyTeleop~\cite{qin2023anyteleop} proposed a solution to the self-occlusion problem by integrating images from multiple cameras, each offering different perspectives.
To further enhance precision in observation, ACE~\cite{yang2024ace} mounted the camera under the end-effector of the exoskeleton to maintain a clear view of the hands and wrists.
MimicPlay~\cite{wang2023mimicplay} employed two calibrated cameras in different viewpoints to reconstruct 3D hand locations. The teleoperation data for the robot was collected using the RoboTurk system~\cite{mandlekar2018roboturk}, which operated via an IMU-equipped smartphone.

\subsubsection{Mocap Gloves}
Motion capture systems typically utilize stable hardware devices such as multi-camera setups with markers, IMU sensors, and RGB-D cameras. These devices are robust against changes in lighting, occlusion, and complex backgrounds. Mocap gloves collect human hand motion data directly via sensors, enabling ideal real-time performance and significantly improving data collection efficiency in teleoperation. Although motion capture gloves are expensive, they provide precise hand tracking \cite{caeiro2021systematic}.
Wang et al. \cite{wang2024dex} introduced DexCap, a portable motion capture system. It includes a mocap glove for accurate finger joint tracking, a single-view camera for 6-DoF wrist pose tracking, and an RGB-D LiDAR camera for observing the surrounding 3D environment. With this precise 3D hand motion data, the proposed DexIL system can effectively learn bimanual dexterous manipulation skills. The remote system features two Franka Emika robotic arms, each with a LEAP dexterous robotic hand.
Similarly, Mosbach et al. \cite{mosbach2022accelerating} used the SenseGlove DK1, a force-feedback glove, to capture hand joint movements, with hand tracking facilitated by a camera mounted on a headset.

\subsubsection{VR/AR Controllers}
VR devices typically include a head-mounted display, a tracking system, and input devices. The head-mounted display provides an immersive visual experience with high-resolution screens and head motion tracking. The tracking system captures the user's movements to ensure that interactions in the virtual environment correspond to real-world actions. Input devices, such as controllers or gloves, facilitate user interaction within the virtual space.
Zhang et al. \cite{zhang2018deep} developed a teleoperation system using consumer-grade VR devices to control a PR2 robot. Following this, methods utilizing low-cost equipment \cite{arunachalam2023holo,radosavovic2023real} demonstrated high-quality teleoperation through mixed reality. To streamline scene construction, Mosbach et al. \cite{mosbach2022accelerating} explored VR teleoperation in simulated environments for manipulation tasks.
Recently, Bunny-VisionPro \cite{ding2024bunny} equipped the Apple VisionPro with a haptic module to provide tactile feedback. Similarly, Open-television \cite{cheng2024opentelevision} used an active camera mounted on a humanoid robot to capture first-person stereo videos. This approach enhances the robot's ability to perform precise and context-aware actions by providing a dynamic, real-time visual perspective similar to human vision.
Lin et al. \cite{lin2024learning} introduced a low-cost teleoperation system, HATO, combining two Psionic Ability Hands for prosthetic use with UR5e robot arms. The system utilizes two Meta Quest 2 VR controllers with IMU sensors to capture hand spatial positions and orientations, translating controller inputs into multi-fingered hand poses.

\subsubsection{Exoskeleton and Bilateral Systems}

The majority of the aforementioned methods focus on manipulating the robot's end-effector in task space in Cartesian coordinates.
While setting the robot's end-effector position is convenient, it has drawbacks. For robots with multiple DoF, computationally demanding inverse kinematics (IK) calculations are required, which can be problematic in real-time control scenarios. These complexities may cause response delays and compromise operational precision. Furthermore, singularities in the motion trajectory may lead to indeterminate or nonexistent IK solutions, resulting in control failures. In the following sections, we discuss studies that aim to synchronize human and robot movements in the joint space.

Exoskeletons are wearable devices that gather and analyze user motion data. Fabian \cite{falck2019vito} developed a lightweight exoskeleton, DE VITO, for measuring human arm movements to teleoperate the mobile robot DE NIRO\cite{falck2018human}. AirExo \cite{fang2024airexo} presented a framework for whole-arm dexterous manipulation adaptable to different robot arms using interchangeable 3D-printed components for robots divergent in morphology.

\begin{table*}[t]
\centering
\renewcommand{\arraystretch}{1.5}
\setlength{\tabcolsep}{5pt}

\arrayrulecolor{black} 
{\color{black} 
\caption{Comparison of Video-based Imitation Learning Approaches}
\label{tab:video_imitation_methods}

\begin{tabular}{>{\centering\arraybackslash}m{4cm}|m{3.2cm}|m{4.5cm}|m{5cm}}
\hline
\raisebox{0.6ex}{\textbf{Method Category}} \rule{0pt}{4ex} & 
\raisebox{0.6ex}{\textbf{Representative Method}} \rule{0pt}{4ex} & 
\raisebox{0.6ex}{\textbf{Data Source}} \rule{0pt}{4ex} & 
\raisebox{0.6ex}{\textbf{Key Characteristics}} \rule{0pt}{4ex} \\
\hline

\textbf{Motion-centric Imitation Learning} &
\begin{itemize}[nosep,leftmargin=3mm]
\vspace{4pt}
\item DexMV \cite{qin2022dexmv}
\item Robotic Telekinesis \cite{sivakumar2022robotic}
\item Track2Act \cite{bharadhwaj2024track2act}
\end{itemize} &
\begin{itemize}[nosep,leftmargin=3mm]
\vspace{4pt}
\item Unlabeled third-person real-world videos
\item Unlabeled third-person unpaired videos
\item Unlabeled videos, limited real-world data
\end{itemize} &
\begin{itemize}[nosep,leftmargin=3mm]
\vspace{4pt}
\item Reconstructs 3D hand-object trajectories or end-effector trajectories from videos and maps them to robot actions, enabling cross-domain imitation and improved generalization.
\end{itemize} \\
\hline

\textbf{Synthetic Video for Policy Learning} &
\begin{itemize}[nosep,leftmargin=3mm]
\vspace{4pt}
\item Gen2Act \cite{bharadhwaj2024gen2act}
\item NIL \cite{albaba2025nil}
\vspace{-8pt}
\end{itemize} &
\begin{itemize}[nosep,leftmargin=3mm]
\vspace{12pt}
\item Text-generated unlabeled synthetic videos
\item Diffusion-generated unlabeled visual videos
\end{itemize} &
\begin{itemize}[nosep,leftmargin=3mm]
\vspace{4pt}
\item Uses language-conditioned video generation to synthesize training demonstrations, removing the need for expert data.
\item Learns policies via perceptual similarity in generated videos, supporting large-scale training and generalization.
\end{itemize} \\
\hline

\textbf{Representation Learning for Generalization} &
\begin{itemize}[nosep,leftmargin=3mm]
\vspace{4pt}
\item Ag2Manip \cite{li2024ag2manip}
\end{itemize} &
\begin{itemize}[nosep,leftmargin=3mm]
\vspace{4pt}
\item Unlabeled human action videos
\end{itemize} &
\begin{itemize}[nosep,leftmargin=3mm]
\vspace{4pt}
\item Learns agent-agnostic action embeddings to support imitation across different robot embodiments and improve generalization.
\end{itemize} \\
\hline


\textbf{Task-specific Architectures and Learning Objectives} &
\begin{itemize}[nosep,leftmargin=3mm]
\vspace{4pt}
\item Bi-KVIL \cite{gao2024bi}
\item Rank2Reward \cite{yang2024rank2reward}
\item ViViDex \cite{chen2024vividex}
\end{itemize} &
\begin{itemize}[nosep,leftmargin=3mm]
\vspace{4pt}
\item Unlabeled third-person dual-hand videos
\item Unlabeled videos
\item Unlabeled noisy videos with no ground-truth state
\end{itemize} &
\begin{itemize}[nosep,leftmargin=3mm]
\vspace{4pt}
\item Specialized frameworks modeling bimanual coordination, reward learning without labels, and hierarchical learning architectures for robust imitation.
\end{itemize} \\
\hline

\end{tabular}

} 
\arrayrulecolor{black} 

\vspace{0.1in}

\end{table*}

Another approach involves a bilateral framework, where the movements of the leader robot are mirrored by the follower robot. Any resistance or force encountered by the follower is communicated back to the leader, enabling precision and tactile sensation tasks.
Kim et al.~\cite{kim2023training} developed a controller with Denavit-Hartenberg parameters matching the teleoperated dual-arm robot alongside a calibration method to reduce gravitational errors. For demonstrations without real robots, the controller uses force/torque sensors identical to those in real robots to provide force feedback.
Recently, employing cost-effective arms with comparable size as leaders and followers, ALOHA \cite{zhao2023learning} utilized structurally analogous robotic arms with identical joint spaces for teleoperation.
Expanding upon this concept, Mobile ALOHA \cite{fu2024mobile} integrated the system with an automated guided vehicle to establish a whole-body teleoperation system. 
GELLO \cite{wu2024gello} reduced costs by replacing the real robotic arms at the local site with scaled kinematically equivalent 3D-printed parts, achieving one-to-one joint mapping. After that, UMI \cite{chi2024universal} further eliminated the need for physical robot arms by using hand-held grippers, providing a more portable interface for in-the-wild data collection.

Differing from specific robot methods,
AnyTeleop \cite{qin2023anyteleop} introduced a unified system supporting multiple robot arms and dexterous hands through a general human-robot hand retargeting method.
This system supports different arms by generating trajectories based on the estimated Cartesian end-effector pose. 
ACE \cite{yang2024ace} developed a cross-platform visual-exoskeleton teleoperation system compatible with diverse robot hardware, including various end-effectors such as grippers and multi-finger hands, offering flexibility. Its exoskeleton arm features high-resolution encoders for precise joint position readings, ensuring accurate end-effector tracking.

\color{black}
Assessing teleoperation interfaces requires considering both interface fidelity and the quality of collected demonstrations. Key evaluation criteria include spatial and temporal tracking precision, the consistency between operator commands and robot execution under communication latency, control bandwidth, and the resolution of feedback. Demonstration quality is typically reflected in task success rate, trajectory smoothness, and the frequency of effective error recovery during operation. An important downstream measure is the interface-to-policy transfer efficiency, which captures how interface characteristics influence the performance and generalization of imitation learning policies. Interfaces with high fidelity and low latency generally enable richer demonstrations and yield superior policy performance in fine-grained manipulation tasks, while lower-bandwidth and less complex interfaces may be sufficient for coarse and high-level motion primitives. Incorporating these metrics together with empirical evidence provides a principled basis for interface comparison, supporting data collection strategies that balance hardware complexity, operator skill, and final policy effectiveness.
\color{black}

%

%

\begin{table*}[t]
\centering
\renewcommand{\arraystretch}{1.5}
\setlength{\tabcolsep}{5pt}

\arrayrulecolor{black}
{\color{black}

\caption{Categorization of Benchmark Datasets for Robotic Imitation Learning}
\label{tab:imitation_learning_dataset_categories}

\begin{tabular}{>{\centering\arraybackslash}m{4cm}|m{3cm}|m{5cm}|m{5cm}}
\hline
\textbf{Method Category} & \textbf{Representative Datasets} & \textbf{Scale} & \textbf{Sensory Modalities} \\
\hline

\textbf{Human-teleoperated Datasets} & 
\begin{itemize}[leftmargin=3mm, rightmargin=0mm]
\vspace{2pt}
\item MIME \cite{sharma_multiple_2018}
\item RH20T \cite{fang_rh20t_2024}
\item BridgeData \cite{ebert_bridge_2022}
\item BridgeData V2 \cite{walke2023bridgedata}
\end{itemize} &
\begin{itemize}[leftmargin=3mm, rightmargin=0mm]
\vspace{2pt}
\item 8,260 demos, 20 tasks
\item 110K+ multimodal sequences
\item 7,200 demos, 71 tasks, 10 domains
\item 60,096 trajectories, 24 environments
\end{itemize} &
\begin{itemize}[leftmargin=3mm, rightmargin=0mm]
\vspace{2pt}
\item RGB-D, Kinesthetic trajectories
\item RGB-D, Audio, Tactile, Proprioception
\item Egocentric vision, Action sequences
\item Vision, Language embeddings
\end{itemize} \\
\hline

\textbf{Augmented Datasets} &
\begin{itemize}[leftmargin=3mm, rightmargin=0mm]
\vspace{2pt}
\item RoboAgent \cite{bharadhwaj2024roboagent}
\end{itemize} &
\begin{itemize}[leftmargin=3mm, rightmargin=0mm]
\vspace{2pt}
\item 7,500 → 98K* augmented trajectories
\end{itemize} &
\begin{itemize}[leftmargin=3mm, rightmargin=0mm]
\vspace{2pt}
\item RGB, Semantic masks
\end{itemize} \\
\hline

\textbf{Synthetic Datasets} &
\begin{itemize}[leftmargin=3mm, rightmargin=0mm]
\vspace{2pt}
\item MimicGen \cite{mandlekar2023mimicgen}
\end{itemize} &
\begin{itemize}[leftmargin=3mm, rightmargin=0mm]
\vspace{2pt}
\item 200 → 50K synthesized demos, 18 tasks
\end{itemize} &
\begin{itemize}[leftmargin=3mm, rightmargin=0mm]
\vspace{2pt}
\item Object poses, Motion primitives
\end{itemize} \\
\hline

\textbf{Dexterous and Bimanual Manipulation Datasets} &
\begin{itemize}[leftmargin=3mm, rightmargin=0mm]
\item ARCTIC \cite{fan2023arctic}
\item DexGraspNet \cite{wang2023dexgraspnet}
\item OAKINK2 \cite{2403.19417}
\end{itemize} &
\begin{itemize}[leftmargin=3mm, rightmargin=0mm]
\item 2.1M frames (3D meshes)
\item 1.32M grasps, 5,355 objects
\item 627 sequences, 4.01M frames
\end{itemize} &
\begin{itemize}[leftmargin=3mm, rightmargin=0mm]
\vspace{2pt}
\item Multi-view RGB, Contact dynamics
\item Tactile sensing, Physics simulation
\item Multi-view capture, 3D pose annotations
\end{itemize} \\
\hline

\end{tabular}


\begin{flushleft}
\footnotesize *RoboAgent employs semantic-preserving augmentation techniques to scale demonstrations
\end{flushleft}

}
\arrayrulecolor{black}

\end{table*}

\color{black}

\subsection{Learning from Video Demonstrations for Dexterous Manipulation}

Video-based demonstration learning offers a promising solution to the challenge of acquiring robotic manipulation skills in the absence of expensive expert demonstrations. Table~\ref{tab:video_imitation_methods} presents a summary of representative methods in this paradigm. By extracting structured behavioral cues from a range of human videos including recorded, synthetic, and web-sourced data, these methods transform raw visual observations into robot-interpretable representations, facilitating robust generalization across diverse tasks, objects, and embodiments.

\subsubsection{Motion-centric Imitation Learning}
A major line of work adopt motion-centric imitation, where human hand–object interactions were converted into robot-executable trajectories. Qin et al. \cite{qin2022dexmv} proposed a method that reconstructed 3D hand–object poses from third-person human videos and re-targeted them to dexterous robot hardware, enabling high-fidelity policy learning without the need for direct teleoperation or egocentric demonstrations. Building upon this approach, Bharadhwaj et al. \cite{bharadhwaj2024track2act} predicted object-relative end-effector trajectories from unstructured videos and fine-tune residual policies on robots, thereby achieving generalization across novel scenes and object configurations with minimal real-world training. Complementary to these approaches, Sivakumar et al. \cite{sivakumar2022robotic} directly mapped unstructured human hand motions from third-person videos to dexterous robotic hands, supporting real-time teleoperation without paired demonstrations.

\subsubsection{Synthetic Video-driven Imitation Learning}
Another direction leverage synthetic video as training signals for policy learning. Bharadhwaj et al. \cite{bharadhwaj2024gen2act} proposed a method that generated realistic manipulation videos from textual prompts and internet instructions, which were then employed to train robot policies capable of generalizing to unseen categories and actions. Building upon this approach, Albaba et al. \cite{albaba2025nil} introduced a diffusion-based video generation framework that replaced explicit action labels with perceptual similarity, providing a scalable pathway for training robots with diverse morphologies. By eliminating the reliance on real or expert demonstrations, these approaches significantly expanded the scalability of video-based demonstration learning.

\subsubsection{Representation Learning for Generalization}
A third stream of research focuses on representation learning for generalization, aiming to decouple policy learning from specific tasks and embodiments. Singh et al. \cite{singh2024hand} proposed a method that spatially aligned human hands and objects in 3D, producing consistent motion patterns that could pretrain general-purpose manipulation policies and enable task-agnostic skill acquisition. Building upon this idea, Li et al. \cite{li2024ag2manip} learned shared agent-agnostic visual–action embeddings from diverse human operation videos, achieving robust sim-to-real transfer even on previously unseen tasks and object configurations. These approaches highlighted the potential of abstracting beyond raw trajectory reconstruction to acquire transferable skills across domains.

\begin{table*}[ht]
\centering
\renewcommand{\arraystretch}{1.3}
\setlength{\tabcolsep}{8pt}
\color{black}
\caption{Rubric for Imitation Learning Dataset Quality Assessment}
\label{tab:dataset_rubric}
\begin{tabular}{p{0.18\linewidth} p{0.3\linewidth} p{0.4\linewidth}}
\hline
\textbf{Criterion} & \textbf{Description} & \textbf{Example Indicators} \\
\hline
Sensor Modality Richness & Extent and variety of sensing channels provided. & RGB, depth, tactile, proprioception, force--torque, audio \\
Annotation Quality & Accuracy, and consistency of labels across modalities and tasks. & Object pose precision, end-effector position, task boundary clarity \\
Task \& Scene Diversity & Breadth of tasks, objects, and environmental variations covered. & Number of object categories, scene layouts, lighting conditions \\
Physical Realism & Degree to which simulated/augmented data approximates real-world physics and appearance. & Visual fidelity, dynamics accuracy, sim-to-real transfer success \\
\hline
\end{tabular}
\end{table*}

\subsubsection{Task-specific Architectures and Objectives}
Finally, several task-specific architectures and learning objectives have been designed to address unique challenges in video-based imitation. Gao et al. \cite{gao2024bi} introduced a hybrid master–slave modeling strategy with low-level geometric constraints for dual-hand coordination, enabling robust reproduction of bimanual manipulation in cluttered environments. Building upon this idea, Shaw et al. \cite{shaw2023videodex} extracted vision–action–force relations from human demonstrations to provide physical and semantic priors for dexterous robot learning, reducing the dependence on paired demonstrations. Extending this approach, Chen et al. \cite{chen2024vividex} presented a three-stage framework that combined video demonstrations, reinforcement learning-based trajectory optimization, and vision-only policy learning without privileged object information, thus enabling physically executable policies from noisy demonstrations. Yang et al. \cite{yang2024rank2reward} addressed the reward specification bottleneck by constructing progress-based rewards from temporal rankings of video frames, supporting both reinforcement and adversarial imitation without access to action annotations. Collectively, these approaches advanced the practicality of video-based demonstration learning by tackling challenges such as bimanual coordination, noisy inputs, reward learning, and task-specific constraints.

\color{black}

\color{black}

\subsection{Datasets and Benchmarks}

\subsubsection{Human-teleoperated Datasets}
A straightforward approach to collecting data for IL is through human teleoperation, where operators directly control robots to perform tasks. MIME \cite{sharma_multiple_2018} represented an early large-scale effort in this direction, providing 8,260 human–robot demonstrations across 20 diverse tasks, along with both human demonstration videos and kinesthetic robot trajectories. RH20T \cite{fang_rh20t_2024} extended this paradigm by scaling up to over 110,000 multimodal manipulation sequences and incorporating richer sensing modalities—visual, tactile, audio, and proprioceptive—captured through teleoperation interfaces with force–torque sensors and haptic feedback. This multimodal design enabled one-shot IL across a wider range of tasks, robots, and environments. Building on the same teleoperation foundation, BridgeData \cite{ebert_bridge_2022} focused on cross-environment generalization, offering 7,200 demonstrations across 71 tasks in 10 environments, primarily in kitchen settings. Its successor, BridgeData V2 \cite{walke2023bridgedata} expanded the dataset to 60,096 trajectories in 24 environments, covering tasks from pick-and-place to complex manipulations, and supporting scalable robot learning with multi-task and language-conditioned objectives.

\subsubsection{Augmented Datasets}
To overcome the time-consuming and labor-intensive process of collecting many human demonstration data for IL, some datasets leverage data augmentation to expand existing demonstrations. RoboAgent \cite{bharadhwaj2024roboagent} began with 7,500 teleoperated trajectories and scaled them up to roughly 98,000 using semantic augmentations, eliminating the need for additional human or robot effort. CyberDemo \cite{wang2024cyberDemo} adopted a similar strategy, collecting demonstrations in both simulated and real environments before applying extensive augmentations that introduced visual and physical variations, improving policy robustness and generalization.

\subsubsection{Synthetic Datasets}
Some datasets employ simulated or synthetic demonstration generation systems to expand the amount of training data from a small set of human demonstrations. MimicGen \cite{mandlekar2023mimicgen} generated over 50,000 demonstrations across 18 tasks from roughly 200 human examples by adapting object-centric manipulation behaviors to new contexts via trajectory transformations. IntervenGen \cite{2405.01472} built on this idea by autonomously producing large sets of corrective interventions from minimal human input, thereby improving policy robustness to distribution shifts. DiffGen \cite{2405.07309} adopted a similar strategy but incorporated differentiable physics simulation, photorealistic rendering, and vision–language models to create realistic robot demonstrations directly from text instructions.

\subsubsection{Dexterous and Bimanual Manipulation Datasets}
Focusing on dexterous manipulation and hand–object interactions, ARCTIC \cite{fan2023arctic} offered 2.1 million videos with precise 3D hand–object meshes and dynamic contact data, enabling the study of bimanual manipulation of articulated objects. DexGraspNet \cite{wang2023dexgraspnet} provided 1.32 million grasps for 5,355 objects using ShadowHand, with each grasp physically validated in simulation to ensure stability. OAKINK2 \cite{2403.19417} extended the scope to real-world bimanual object manipulation, comprising 627 sequences and 4.01 million frames from multi-view captures with detailed pose annotations for human bodies, hands, and objects.
\color{black}
Table \ref{tab:imitation_learning_dataset_categories} presents a concise summary of the representative benchmarks and datasets discussed above.
\color{black}

\color{black}
From the perspective of dataset usability, several interdependent factors critically shape the practical value of an imitation learning dataset (as shown in Table~\ref{tab:dataset_rubric}). Foremost among these is sensor modality richness: datasets offering more comprehensive multi-modal signals (e.g., depth image, tactile feedback, proprioceptive states, force–torque readings, audio) generally deliver superior informational content and flexibility compared to RGB-only collections. Such multi-modal data facilitates more accurate perception, reasoning over physical contacts, and cross-modal policy learning. Annotation quality is another decisive factor; precise labels for object poses, manipulator or end effector positions, and task boundaries directly constrain the attainable performance of learned models. Task and scene diversity further influences generalization: datasets encompassing a wide spectrum of object categories, environmental configurations, and illumination conditions are better poised to support robust and transferable policies than those restricted to narrow, repetitive settings. For simulated or augmented datasets, physical realism is especially critical, as greater realism narrows the sim-to-real gap and boosts the likelihood that policies trained virtually will operate successfully on physical robots. 
\color{black}

\section{Challenges and Future Directions in Imitation Learning-Based Dexterous Manipulation}

\begin{figure}[t]
    \centering
    \includegraphics[width=1\linewidth]{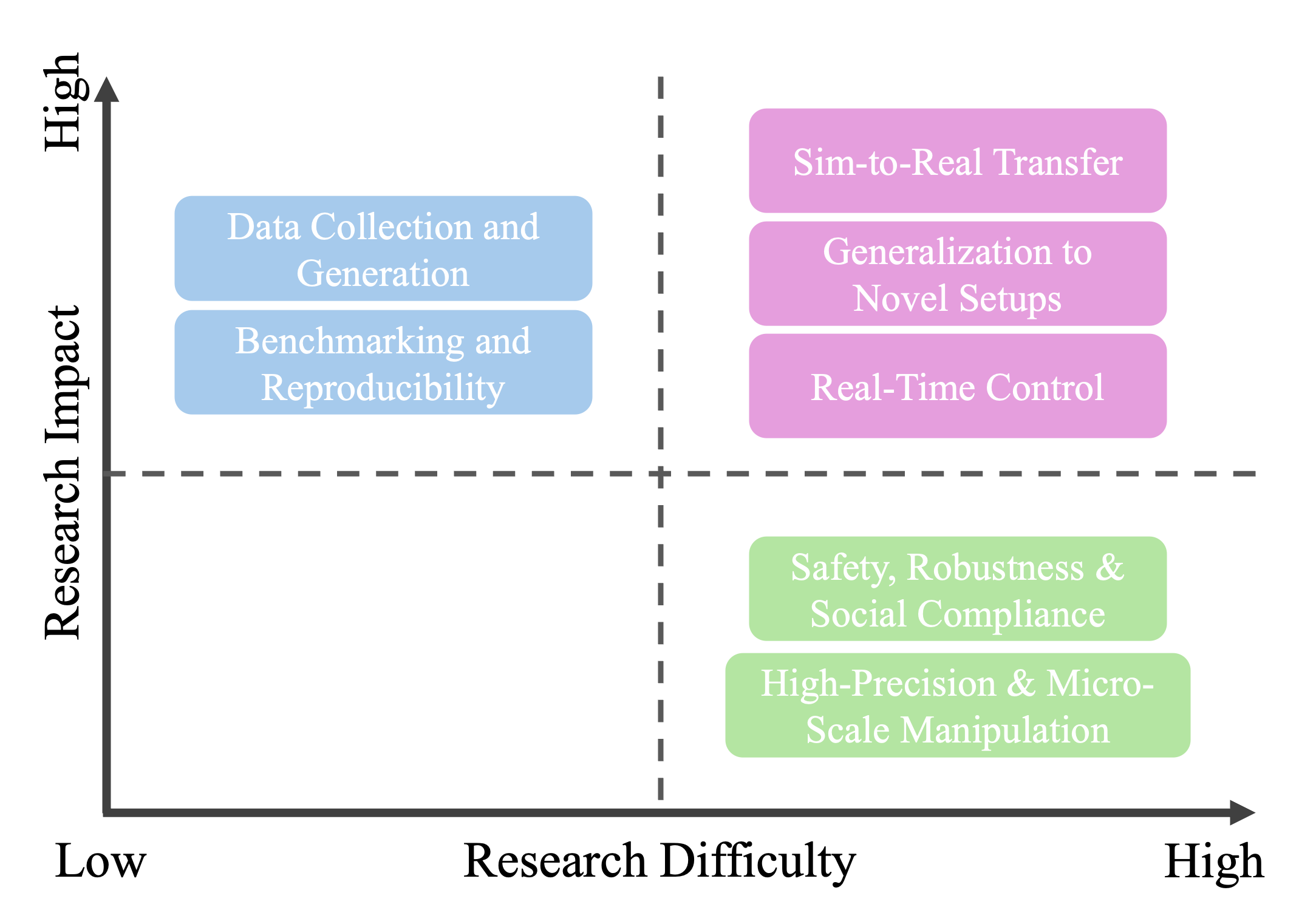}
      \vspace{-0.3in}
  \caption{Prioritization matrix of key challenges in imitation learning-based dexterous manipulation.}
  \vspace{-0.1in}
\label{fig:prioritization_matrix}
\end{figure}

IL-based dexterous manipulation poses unique challenges due to the inherent complexities of both IL and dexterous control. Despite significant advancements over the past decade, several challenges hinder its human-level dexterity and real-world applicability. Fig.~\ref{fig:prioritization_matrix} summarizes the challenges based on research impact and technical difficulty, highlighting immediate priorities further discussed in this section.

\subsection{Data Collection and Generation}

Data collection and generation for imitation learning-based dexterous manipulation pose several challenges, including heterogeneous data fusion, data diversity, high-dimensional data sparsity, and data collection costs.

\subsubsection{Heterogeneous Data Fusion} Dexterous manipulation relies on multi-modal sensory inputs (e.g., visual, tactile, proprioceptive, and force), each with varying sampling rates, noise characteristics, and spatial-temporal resolutions, making data integration and synchronization challenging. Moreover, differences in embodiments and gripper designs introduce additional complexities. For example, demonstrations collected with one robotic hand may not directly generalize well to another due to variations in kinematics, actuation mechanisms, and sensor placements. Addressing these challenges requires (1) multi-modal alignment techniques to improve sensor fusion and (2) cross-embodiment learning frameworks for better transferability across robotic platforms and embodiments.

\subsubsection{Data Quantity, Quality, and Diversity} Ensuring sufficient data quantity, quality, and diversity is challenging because collecting expert demonstrations for dexterous tasks at scale is labor-intensive and expensive. Even small variations in object properties, task conditions, or environmental factors can significantly affect manipulation policies, making it difficult for IL models to generalize. Future research should explore synthetic data augmentation, domain randomization, and generative models to efficiently generate diverse training datasets. Scalable and automated data collection methods, such as crowdsourced teleoperation, where multiple users remotely control robots to provide varied demonstrations, and self-supervised learning, where robots autonomously collect and label data through interaction and feedback, can further mitigate data collection bottlenecks. Additionally, establishing standardized data collection protocols and defining robust evaluation metrics for data quality and diversity will be essential to ensure consistency and reliability.

\subsubsection{High-Dimensional Data Sparsity}
Data sparsity in high-dimensional action spaces limits the effectiveness of learned policies, as dexterous manipulation requires precise finger coordination, force regulation, and contact-rich interactions that demonstrations alone struggle to capture comprehensively. Hierarchical representation learning can potentially mitigate this challenge by structuring high-dimensional action spaces into more learnable subspaces. In dexterous manipulation, decomposing control policies into hierarchical levels—such as low-level motor commands, mid-level grasp strategies, and high-level task affordances—allows models to extract structured representations, improving learning efficiency and reducing dependence on large-scale demonstrations.

RL fine-tuning further complements IL by refining dexterous manipulation policies beyond demonstrated behaviors. Fine-tuning in simulation enables robots to explore variations in object properties, task conditions, and environmental dynamics that may not be covered in the demonstration data. However, effective sim-to-real transfer techniques and high-fidelity physics engines are crucial to bridging the gap between simulated training and real-world execution.

\subsubsection{Data Collection Costs}
The high cost and complexity of data collection pose barriers to scaling IL for dexterous manipulation. Traditional methods often require specialized motion capture systems, high-precision force sensors, and complex teleoperation setups, which are expensive, labor-intensive, and impractical for large-scale data acquisition. Reducing these barriers requires the development of low-cost, scalable data collection methods, such as wearable sensor systems for capturing human demonstrations and shared autonomy techniques to minimize operator effort. Additionally, establishing standardized data collection protocols and collaborative data-sharing platforms can improve data accessibility and consistency across datasets.

While simulation provides a scalable solution for generating synthetic data in dexterous manipulation, several challenges limit its real-world effectiveness. First, achieving real-world fidelity remains difficult, as physics engines struggle to model contact dynamics, deformable objects, and high-resolution tactile feedback, leading to discrepancies between simulation and reality. Second, ensuring sufficient data diversity is another challenge, as models trained in static or overly idealized environments often fail to generalize to unstructured real-world conditions, and while domain randomization can enhance robustness, excessive variation may reduce learning efficiency or introduce unrealistic artifacts. Third, the sim-to-real gap further complicates deployment, as policies trained in simulation often fail in real-world settings due to sensor noise, unexpected disturbances, and actuation discrepancies. While techniques such as domain adaptation, sim-to-real fine-tuning, and physics-based calibration can help mitigate these challenges, they require substantial computational resources and real-world validation, increasing deployment complexity. 

\subsection{Benchmarking and Reproducibility}

The dependence on real-world hardware experiments and the variability in simulation environments pose significant challenges for benchmarking and reproducibility in imitation learning-based dexterous manipulation. Unlike computer vision or natural language processing, where large-scale datasets enable standardized evaluations, dexterous manipulation involves physical interactions, making consistent replication across research efforts difficult. Hardware dependency is a major obstacle, as reproducing results requires access to the same robotic platform, gripper design, sensor setup, and control software, which is often impractical in real-world experiments due to cost, availability, and proprietary constraints.

Simulation-based benchmarks offer a scalable alternative, but the lack of standardized simulation settings, computing environments, and evaluation protocols in physics-based simulators limits fair comparisons across studies. Variability in physics engine configurations, actuator models, contact dynamics, and material properties further exacerbates inconsistencies, making it difficult to establish reliable performance benchmarks and universally comparable evaluation metrics in dexterous manipulation research. Some studies rely on non-physics-based or simplified simulators, which focus on high-level task planning but neglect low-level contact physics modeling. While these environments provide visual realism and scalable training, they introduce a significant sim-to-real gap, failing to capture key aspects of dexterous manipulation, such as precise force interactions and object deformations.

Addressing these challenges requires standardized benchmarking frameworks and open-source datasets for both simulation and real-world experiments. In simulation, standardization should focus on consistent physics parameterization (e.g., contact dynamics, actuator models, material properties) and common environment representations to minimize discrepancies across different physics engines. For real-world experiments, benchmarks should incorporate multi-modal sensory recordings (e.g., RGB-D, tactile, proprioceptive data) and diverse task demonstrations across various robotic embodiments to ensure broader comparability. Additionally, establishing standard evaluation protocols across hardware platforms and physics-based simulators would enable more reliable performance comparisons across studies.

\subsection{Generalization to Novel Setups}

Generalizing IL-based dexterous manipulation policies is challenging due to task and environment variability, adaptive learning limitations, and cross-embodiment adaptability:

\subsubsection{Task and Environment Variability} Learning-based policies often struggle to extend beyond the specific demonstrations to new conditions. Variations in object shapes, sizes, weights, textures, and dynamic interactions, as well as unforeseen obstacles and workspace changes, can significantly degrade performance. Also, these policies may fail when faced with unseen task configurations that require adaptive behavior beyond the demonstrated distribution.

\subsubsection{Adaptive and Continual Learning Frameworks} Traditional IL models do not adapt to new tasks or environmental changes after training. This limitation leads to rigid behaviors that fail to improve with experience. Continual learning frameworks allow robots to learn incrementally from new data without catastrophic forgetting, while adaptive learning methods such as meta-learning and RL fine-tuning enable policies to generalize to new conditions by leveraging prior experience. Additionally, uncertainty-aware models can dynamically adjust decision-making strategies based on real-time feedback, improving generalization in unstructured settings.


\subsubsection{Cross-Embodiment Adaptability} Variability in robot embodiments, gripper designs, sensor configurations, and actuation dynamics poses significant challenges for generalization. A policy trained on one robotic hand may struggle to transfer to another due to differences in degrees of freedom, joint limits, contact dynamics, and control strategies. Even within the same robotic platform, inconsistencies arise from sensor noise, latency, and mechanical tolerances. To address this, morphology-agnostic policy learning can be explored, where models are trained across diverse robotic embodiments to develop transferable representations. Graph-based and latent-space embeddings of robot kinematics could help policies reason about different embodiments more effectively. Additionally, modular policy architectures, where separate components (e.g., perception, control, and adaptation modules) are fine-tuned independently, may enhance transferability. Another promising direction is meta-learning and few-shot adaptation, enabling robots to quickly adjust to new embodiments with minimal data, reducing the need for extensive retraining.

\color{black}
\subsection{Sim-to-Real Transfer}

Simulation provides a scalable and controlled environment for training dexterous manipulation policies. However, the sim-to-real gap remains a major obstacle to deploying IL models on real-world robotic systems. This gap arises from discrepancies in factors such as contact dynamics, sensor noise, actuation delays, and material properties, which are often inadequately modeled in simulation. The following subsection discusses key techniques and ongoing challenges related to sim-to-real transfer.

\subsubsection{Domain Randomization}
One widely adopted strategy for bridging the sim-to-real gap is domain randomization. Instead of replicating real-world conditions exactly, domain randomization introduces variability in simulation parameters. This exposure helps policies generalize to unseen real-world scenarios \cite{tobin2017domain}. However, domain randomization requires manual tuning of parameter distributions, which demands expert knowledge to balance realism and variability. It also struggles to address unmodeled dynamics, such as non-linear material deformations or high-frequency contact interactions, which are crucial for dexterous manipulation. Therefore, future research could focus on combining domain randomization with complementary sim-to-real techniques, such as real-world fine-tuning or adversarial domain adaptation, to achieve more robust and reliable transfer.

\subsubsection{Feature Alignment}
Another promising line of work focuses on feature alignment, where the objective is to map simulated and real-world sensory inputs into a shared latent feature space. This can be achieved through representation learning techniques such as autoencoders or contrastive learning, which minimize the distributional discrepancy between the domains. Similarly, cross-modal embeddings can be learned to align tactile or proprioceptive features from both domains, which is particularly valuable for dexterous tasks involving high-dimensional state observations. These approaches can complement domain randomization by providing a structured way to reduce domain shift.

\subsubsection{Adversarial Domain Adaptation}
Adversarial techniques have also emerged as a powerful way to minimize sim-to-real discrepancies. This approach has been applied to tasks where direct collection of real-world demonstrations is expensive \cite{tzeng2017adversarial}. In robotic manipulation, adversarial adaptation can be used in combination with IL by first imitating expert demonstrations in simulation and then fine-tuning the learned policy to fool a domain discriminator, ensuring smoother deployment in real-world settings. Future research could explore integrating adversarial adaptation with generative models or self-supervised representation learning to improve robustness and scalability, particularly for high-dimensional sensory inputs such as vision and tactile feedback.

\subsubsection{Hybrid Training and Online Adaptation}
Beyond domain adaptation, hybrid training paradigms are gaining significant traction. In these approaches, policies are first pre-trained in simulation for rapid skill acquisition and subsequently fine-tuned on real-world data using IL, RL, or offline corrections \cite{rajeswaran2018learning}. This strategy combines the efficiency and scalability of simulation with the precision of real-world adaptation, thereby reducing the need for extensive real-world data collection. Additionally, self-supervised real-to-sim refinement, where real-world rollouts are used to iteratively adjust simulation parameters, can improve the fidelity of simulated environments and facilitate bi-directional transfer.
\color{black}

\subsection{Real-Time Control}

Dexterous manipulation presents significant computational challenges due to its high-dimensional action spaces and complex dynamics. Achieving real-time execution demands a delicate balance between accuracy and efficiency in terms of both software and hardware.

Efficient real-time control relies on algorithms capable of handling nonlinearities, contact dynamics, and feedback loops while maintaining stability and responsiveness. Model-based approaches, such as optimal control and Model Predictive Control (MPC), leverage system dynamics to generate control policies but often struggle with the complexities of dexterous manipulation. MPC, in particular, provides real-time adaptability through continuous optimization but imposes high computational demands, often requiring specialized hardware acceleration or dedicated edge computing to meet real-time constraints. In contrast, model-free RL learns policies directly from data, bypassing the need for explicit system modeling. While RL offers greater adaptability in high-dimensional, unstructured environments, it remains sample inefficient, prone to slow convergence, and challenging to stabilize, especially for real-time execution. A potential solution is designing hybrid control strategies that combine model-based control for stability with model-free learning for adaptability, improving efficiency without sacrificing robustness. Meanwhile, accelerated learning techniques, such as parallelized RL training and meta-learning, could address sample inefficiency, enabling faster policy convergence.

Hardware architecture is also a key enabler of real-time dexterous manipulation, balancing computational power, latency, and energy efficiency. High-performance computing hardware (e.g., GPUs, TPUs, and FPGAs) is essential for complex model-based and learning-based control strategies but is often constrained by high power consumption and deployment costs. Edge computing and custom ASICs offer low-latency processing but may lack the computational capacity required for large-scale dexterous manipulation policy inference. Cloud computing facilitates large-scale training and high-fidelity simulations; however, real-time reliance on remote processing is limited by communication delays and network instability. Recent advancements in low-power AI accelerators, neuromorphic computing, and distributed edge-cloud architectures have the potential to enhance real-time processing while reducing latency and energy constraints.

\subsection{Safety, Robustness, and Social Compliance}

Ensuring safety, robustness, and social compliance is crucial for real-world dexterous robotics, requiring risk prevention, adaptive error recovery, and human-aware behavior for seamless integration. Real-world dexterous manipulation demands reliable error detection, recovery, and adaptability in dynamic environments. Detecting failures is challenging due to sensor noise, occlusions, and unpredictable interactions, while recovery strategies like re-grasping or trajectory replanning must be executed in real time to maintain stability and task continuity. Future research should address two key aspects. First, large-scale failure datasets and standardized benchmarks are essential for improving data-driven recovery policies. Establishing comprehensive datasets and evaluation protocols for failure detection, uncertainty estimation, and recovery effectiveness would provide a foundation for training and benchmarking robust policies. Second, self-supervised learning for multi-modal anomaly detection could enable robots to autonomously refine their error detection capabilities. By leveraging visual, tactile, and proprioceptive feedback, robots could learn to recognize and anticipate failures in real time, improving adaptability and robustness in dynamic environments.

Safety is equally critical, particularly in real-world deployments where unpredictable interactions and dynamic conditions pose significant risks. In dexterous manipulation, safety considerations involve collision avoidance, force regulation, and compliance control, particularly when interacting with fragile objects or operating near humans. However, achieving these safety measures requires handling varying contact conditions, but sensor noise, occlusions, and data processing delays can reduce reliability. Additionally, while compliant actuators and soft robotic designs help mitigate impact forces, integrating these hardware safety mechanisms involves trade-offs between control precision, responsiveness, and durability. 
Different paradigms also entail distinct safety concerns. End-to-end learning directly maps perception to control, but its black-box nature complicates safety verification and error tracing. Motion module learning decomposes the pipeline into structured stages, improving generalization and safety at the cost of intermediate complexity.

To address safety constraints in IL for real-world dexterous manipulation, recent studies have investigated constrained policy optimization \cite{achiam2017constrained} and Lyapunov-based safe learning \cite{chow2018lyapunov}.
Constrained policy optimization integrates hard limits on torque, contact force, or joint velocity directly into the training objective, often via Lagrangian relaxation or primal-dual methods. For instance, Constrained Trust Region Policy Optimization (C-TRPO) \cite{milosevic2024embedding} augmented the trust-region update rule by incorporating actuator and contact-force limits, ensuring that each policy update remains close to expert demonstrations while strictly respecting physical safety bounds.
In parallel, Lyapunov-based approaches synthesized control policies that provably satisfy safety specifications, constructing Lyapunov functions as certificates of stability~\cite{khansari2011learning,xu2020learning,xu2021learning}. In dexterous manipulation, Neural Lyapunov Control (NLC) \cite{chang2019neural} has been adapted to learn neural network policies that maintain force closure and avoid excessive contact forces.
Recent work combined these paradigms. SafeDiffuser~\cite{xiao2023safediffuser} augmented diffusion-policy-based planning with control barrier functions to enforce real-time safety constraints. By conditioning the diffusion model on expert demonstrations and integrating CBF-based corrections at every denoising step, the method achieves robust generalization to unseen scenarios while guaranteeing safety-critical bounds. 

\textcolor{black}{Beyond technical safety, social compliance is essential for real-world deployment yet remains underexplored, particularly in HRI settings. In the context of dexterous manipulation, prior research has investigated various approaches, such as RL-based methods (e.g. preference-based RL \cite{an2023direct} and inverse RL \cite{das2021model}), MPC frameworks (e.g. socially aware Model Predictive Path Integral control \cite{trevisan2024biased}), and game-theoretic models (e.g. Nash equilibrium formulations \cite{bansal2020bayesian}). However, achieving social compliance remains challenging due to the inherent ambiguity of human comfort, spatial preferences, and social norms, which are often context-dependent and difficult to model explicitly. Additionally, soft social constraints, such as maintaining legibility of motion, avoiding intrusive actions, or signaling intent, are subjective and require learning from diverse human feedback, which is costly and time-consuming. Another challenge lies in the lack of standardized evaluation metrics and benchmarking protocols to assess social compliance. Existing metrics, such as trajectory smoothness, proximity compliance, or motion legibility, are often evaluated qualitatively or through small-scale user studies without a consistent framework. Moreover, there is a lack of publicly available datasets and simulation environments specifically designed for evaluating social compliance in dexterous manipulation, unlike traditional benchmarks for grasping or motion planning (e.g., RLBench~\cite{james2019rlbench} or ManiSkill3~\cite{taomaniskill3}). Without well-defined metrics, datasets, and test scenarios, progress in this area remains fragmented and difficult to measure systematically.}

\color{black}
\subsection{High-Precision and Micro-Scale Manipulation}
While the above discussion focuses on macro-scale object manipulation (e.g., bottles, tools), scaling down to the millimeter or micrometer level (e.g., electronic components, micro-machines) introduces fundamentally different challenges. At these scales, perception limitations, such as low sensor resolution, visual noise, and occlusion, become critical. Actuation requirements also rise sharply, demanding ultra-precise force and motion control within micrometer tolerances to prevent damage to delicate objects. Conventional actuators frequently suffer from backlash, hysteresis, and limited resolution, while achieving high control precision becomes a bottleneck. These challenges have driven the development of advanced control methods \cite{zhong2023spatial, xin2025dynamic, wang2024data}, including force-feedback, adaptive impedance, and hybrid position-force strategies designed specifically for micro-scale tasks.

Moreover, data collection at micro-scales remains a significant challenge. Acquiring large, diverse, and high-quality demonstrations is both costly and time-intensive, given the specialized equipment and delicate experimental setups required. Synthetic data generation through high-fidelity physics-based simulation is a promising direction, but it requires accurate modeling of micro-scale physical effects, such as adhesion forces, surface roughness, or fluid dynamics, which are often negligible in macro-scale tasks. From an algorithmic perspective, recent research \cite{kim2025srt} underscored the potential of hierarchical IL frameworks in addressing dexterous manipulation in micro-scale tasks. These techniques, together with advances in small-scale robots\cite{dong2022untethered}, presented a promising pathway for enabling autonomous micro-manipulation tasks, such as precision assembly and surgical procedures.
\color{black}

\subsection{Roadmap for IL-based Dexterous Manipulation}
To summarize, we propose a potential roadmap to address these issues, distinguishing between low-hanging fruits and grand, long-term challenges.

\subsubsection{Low-Hanging Fruits}
Data collection and benchmarking are high-impact areas that require relatively low research difficulty. Efforts focused on synthetic data generation, domain randomization, and crowdsourced teleoperation can quickly scale up the diversity of training datasets, which is essential for improving the generalization capabilities of models. Furthermore, establishing standardized benchmarking frameworks and evaluation protocols will streamline the process of comparing performance across different studies. These actions will enable faster progress in dexterous manipulation and contribute to broader adoption of the proposed techniques.

\subsubsection{Grand, Long-Term Challenges}
Sim-to-real transfer and real-time control are high-impact challenges that involve significant research difficulty. Bridging the sim-to-real gap will require breakthroughs in domain adaptation, hybrid training paradigms, and better computational resources. These advances are crucial for ensuring the successful deployment of models trained in simulation into real-world environments. Real-time control, particularly for high-dimensional tasks like dexterous manipulation, presents a need for more efficient algorithms, better control strategies, and seamless hardware integration. Solving these challenges will be essential for creating practical, high-performance systems capable of operating in dynamic, unstructured environments.

\subsubsection{Safety and High-Precision Manipulation}
While safety and high-precision manipulation are specialized concerns, they remain essential for real-world applications. These areas are high-difficulty but relatively lower in impact compared to sim-to-real transfer and real-time control. However, safety remains critical for human-robot interaction, especially in unstructured environments, and high-precision control is crucial for tasks like micro-manipulation. Future research should continue to address these challenges for real-world deployment.



\section{Conclusion}
Imitation learning has shown significant promise in enabling robots to perform dexterous manipulation tasks with human-like skill and precision. By learning from human demonstrations, robots can acquire complex manipulation capabilities that are difficult to achieve through traditional programming methods. This survey has provided an overview of the current state-of-the-art in IL-based dexterous manipulation, highlighting key techniques, applications, and challenges.

Despite the progress that has been made, several challenges remain that hinder the practical deployment of these systems. Addressing issues related to data collection, generalization, real-time control, safety, and sim-to-real transfer is essential for advancing the field. Future research should focus on developing optimized IL algorithms, enhancing human-robot collaboration, and integrating advanced sensory systems.

The future of dexterous manipulation holds great potential, with applications ranging from industrial automation to healthcare and service robotics. By continuing to push the boundaries of IL and robotic manipulation, researchers and practitioners can pave the way for more capable, adaptable, and intelligent robots. These advances will not only improve the efficiency and safety of robotic tasks but also open up new possibilities for human-robot collaboration and interaction.


%

\appendices



\section*{Acknowledgment}
We would like to thank Huapeng Li, Qianyi Wang, Mingwu Liu, Shouzheng Wang and Zijing Yang for their contributions to the completion of this survey.


\ifCLASSOPTIONcaptionsoff
  \newpage
\fi



%

\bibliographystyle{IEEEtran} 
\bibliography{references,Yuning_references_update,Yuning_website_references_update} 

%
\vspace{-10pt}
\begin{IEEEbiography}[{\includegraphics[width=1in,height=1.25in,clip,keepaspectratio]{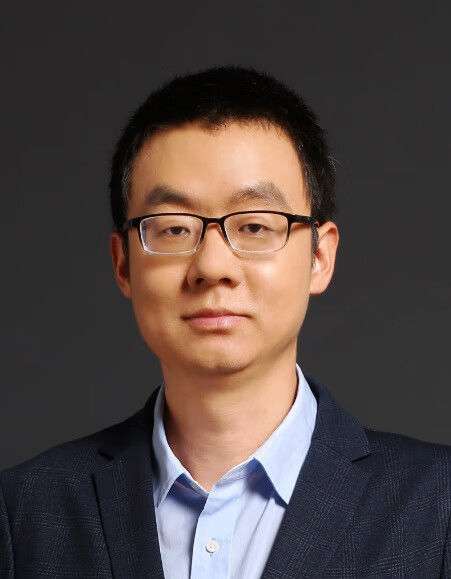}}]{Shan An (Senior Member, IEEE)} received the B.E. degree from Tianjin University in 2007, the M.E. degree from Shandong University in 2010, and the Ph.D. degree in computer science and engineering from Beihang University in 2022. He is currently an Associate Professor with the School of Electrical and Information Engineering, Tianjin University. His research interests include dexterous manipulation, extended reality for robotics, and robotic vision. Prior to his academic career, he accumulated 14 years of extensive industrial experience at the China Academy of Space Technology, Alibaba, and JD Group. He was elected Fellow of the Institution of Engineering and Technology (IET) in 2023 and Fellow of the British Computer Society (BCS) in 2024.
\end{IEEEbiography}
\vspace{-13pt}
\begin{IEEEbiography}[{\includegraphics[width=1in,height=1.25in,clip,keepaspectratio]{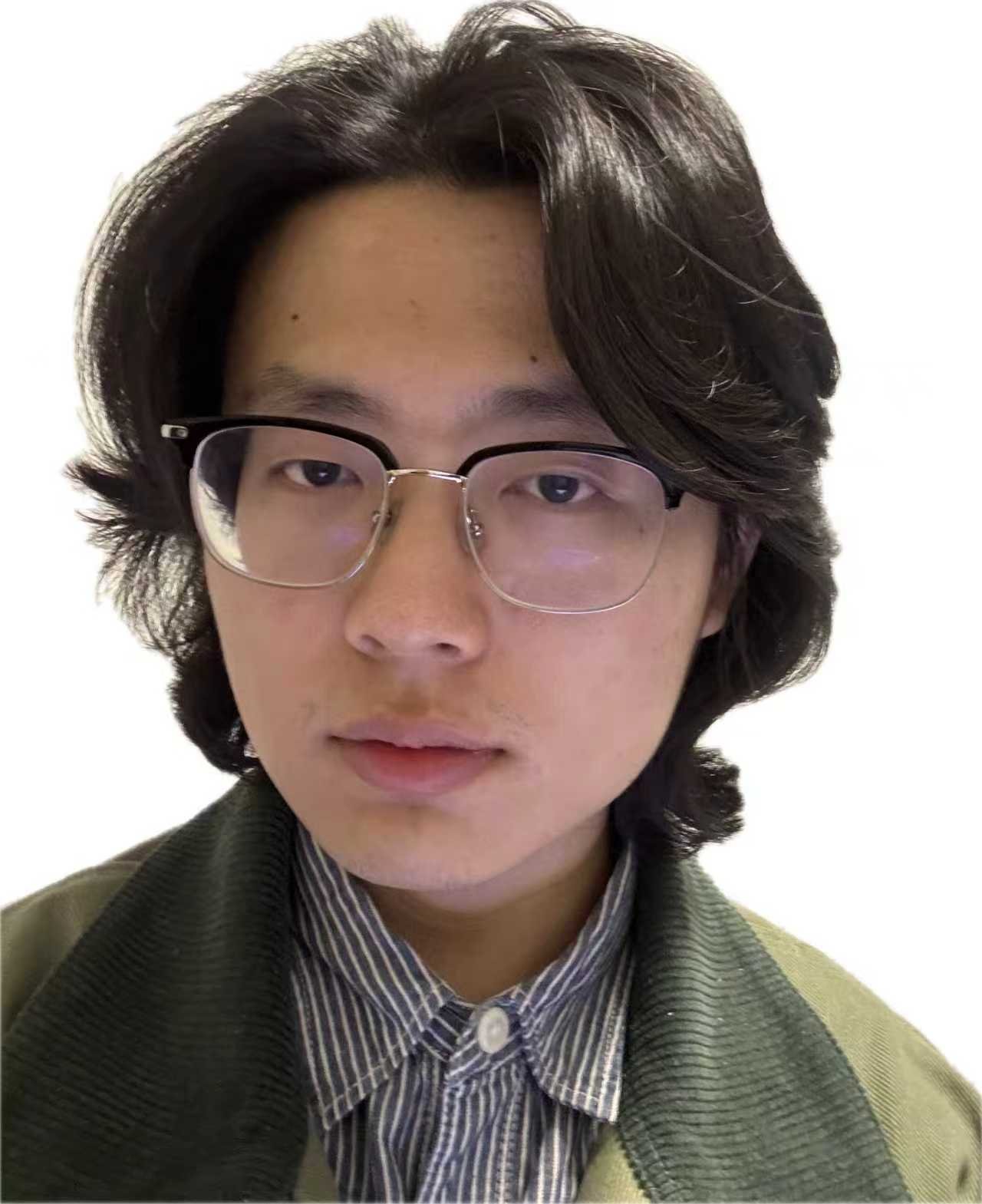}}]
{Ziyu Meng} received the B.S. degree from Shandong University, Jinan, China, in 2022. He is currently pursuing the Ph.D. degree in pattern recognition and intelligent systems at Shandong University, Jinan, China. His current research interests include whole-body control for humanoid robots.
\end{IEEEbiography}

\begin{IEEEbiography}[{\includegraphics[width=1in,height=1.25in,clip,keepaspectratio]{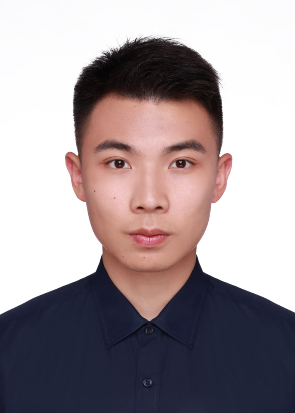}}]{Chao Tang} is currently a postdoctoral researcher at the Division of Robotics, Perception and Learning, KTH Royal Institute of Technology. He received his Ph.D. degree from Southern University of Science and Technology in 2025 and his M.S. degree from the Georgia Institute of Technology in 2020. His research interests include robotic manipulation and grasping, mobile manipulation, human–robot interaction, and semantic reasoning for robots.
\end{IEEEbiography}
\vspace{-13pt}
\begin{IEEEbiography}[{\includegraphics[width=1in,height=1.25in,clip,keepaspectratio]{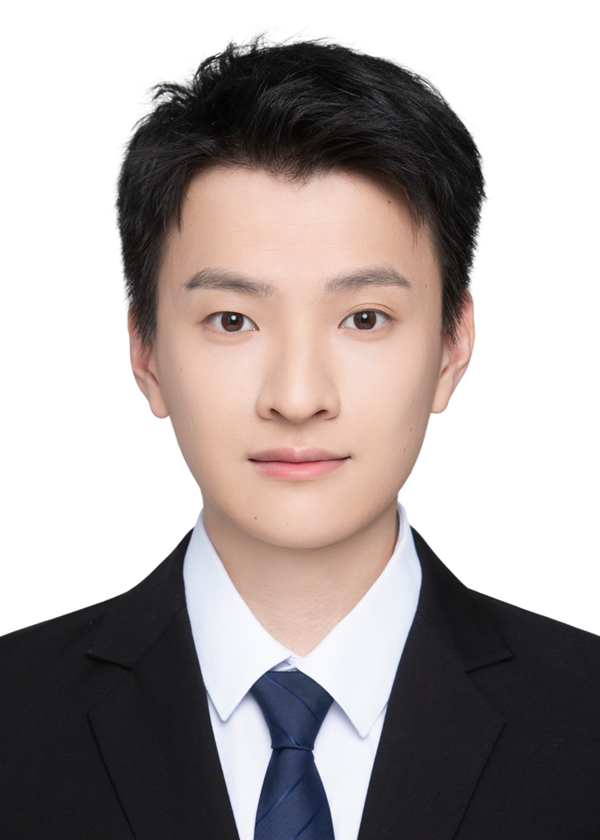}}]{Yuning Zhou} received the B.E. degree in Mechanical Engineering from the University of Leeds and Southwest Jiaotong University (dual-degree program) in 2022, and the M.Sc. degree in Robotics, Systems and Control from ETH Zurich in 2025. His research interests include the design, simulation, and benchmarking of anthropomorphic robotic hands.
\end{IEEEbiography}
\vspace{-13pt}
\begin{IEEEbiography}[{\includegraphics[width=1in,height=1.25in,clip,keepaspectratio]{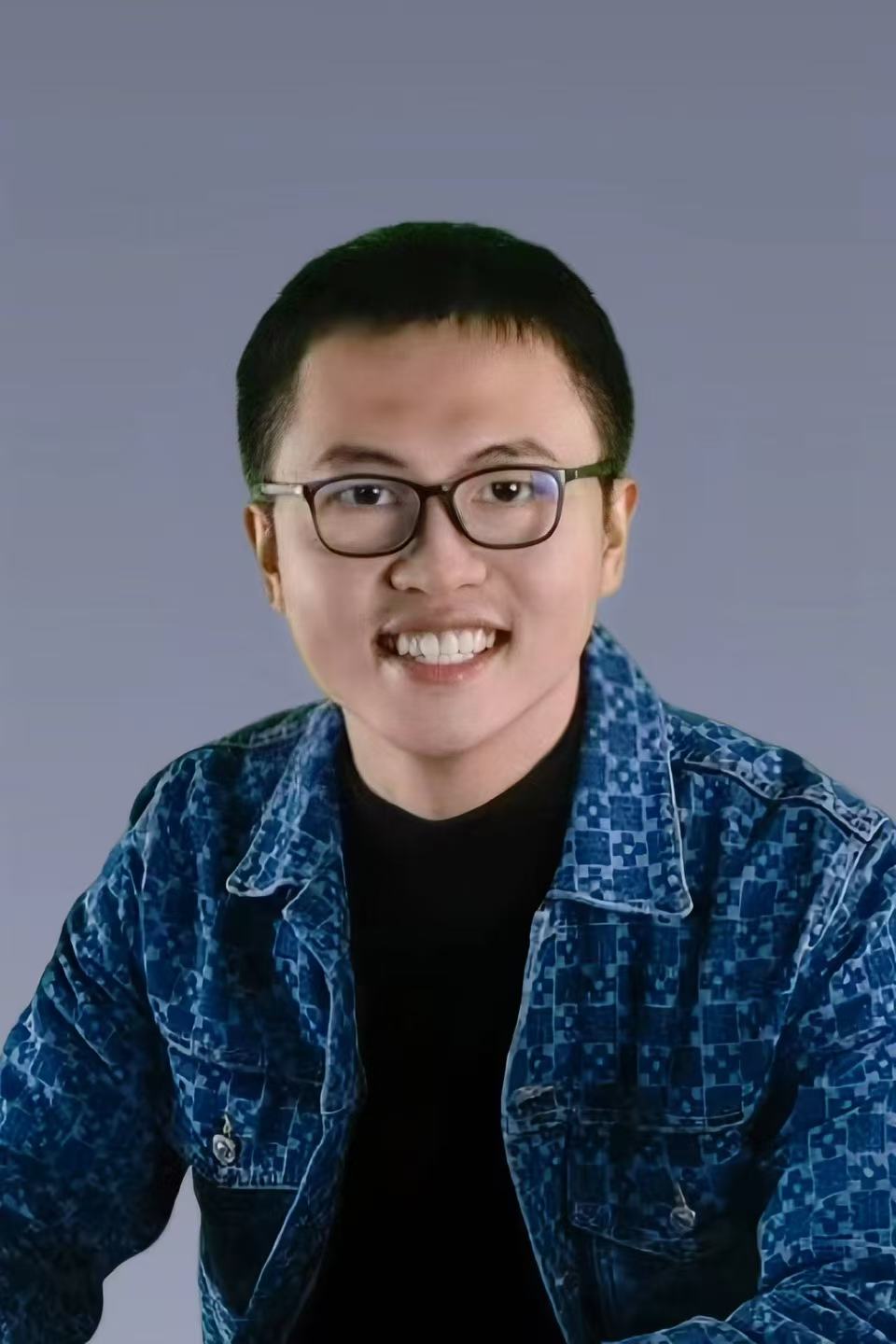}}]{Tengyu Liu} received the B.S. degree in computer science from the University of Illinois Urbana-Champaign (UIUC), Champaign, IL, USA, and the M.S. and Ph.D. degrees in computer science from the University of California, Los Angeles (UCLA), Los Angeles, CA, USA, in 2021. He is currently a Senior Research Scientist at Beijing Institute of General Artificial Intelligence (BIGAI), Beijing, China. His research interests lie at the intersection of 3D computer vision, computer graphics, and robotics, specifically focusing on generalizable dexterous grasping, manipulation, and whole-body control of humanoid and quadruped robots.
\end{IEEEbiography}
\vspace{-13pt}
\begin{IEEEbiography}[{\includegraphics[width=1in,height=1.25in,clip,keepaspectratio]{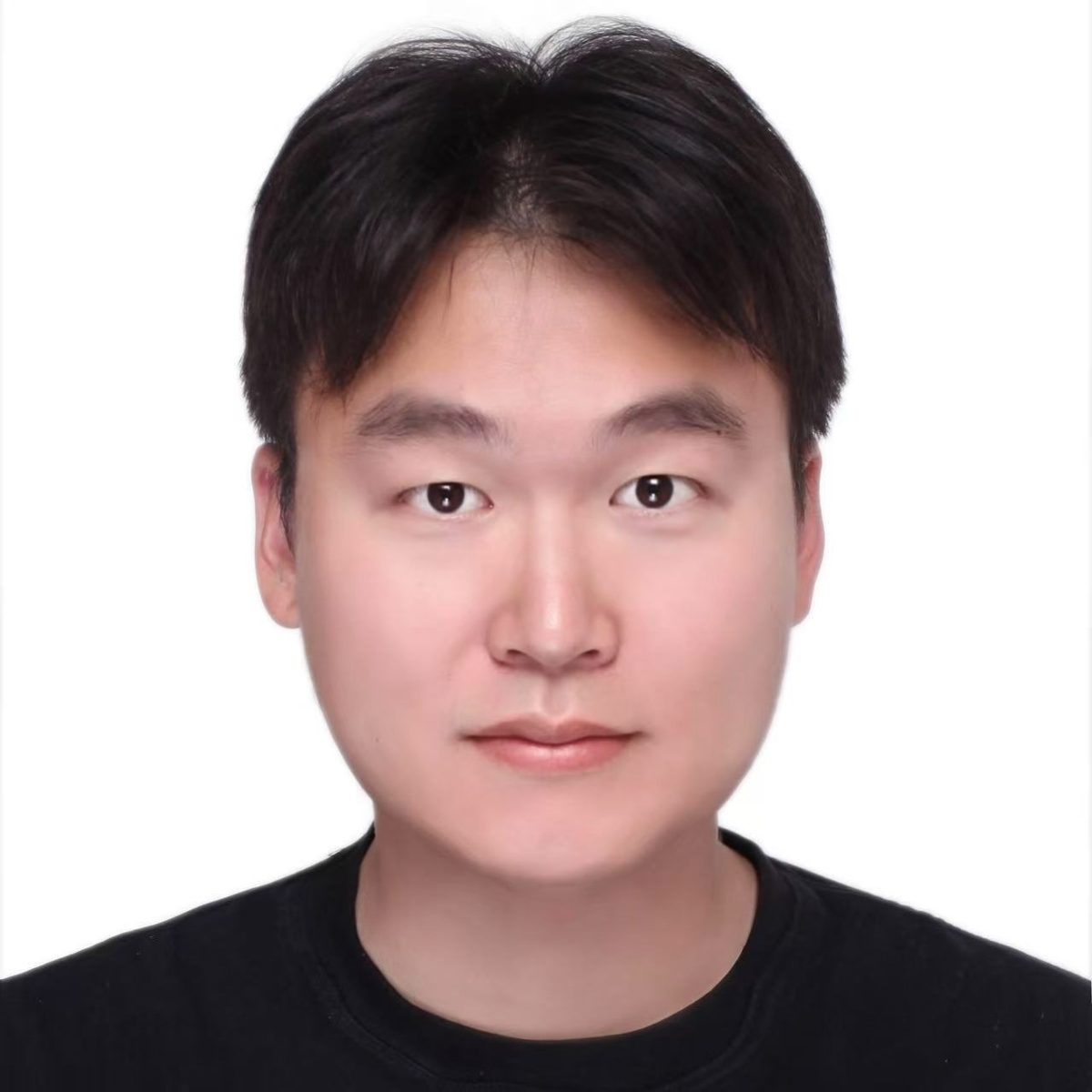}}]{Fangqiang Ding} is a Postdoctoral Associate with the Department of Mechanical Engineering at the Massachusetts Institute of Technology (MIT), Cambridge, MA, USA, working with Dr. Hermano Igo Krebs in The 77 Lab. He was previously a Postdoctoral Fellow at Technion – Israel Institute of Technology, working with Dr. Or Litany. He received the Ph.D. degree in Robotics and Autonomous Systems from the School of Informatics, University of Edinburgh, U.K., and the B.E. degree from Tongji University, Shanghai, China. He was selected as an RSS Pioneer in 2025 for his work on robust spatial perception with 4D radar for mobile autonomy. My research centers on reliable and affordable Physical AI, enabling AI-integrated physical systems (e.g., autonomous vehicles, robots, and IoT) to operate responsibly around humans and deliver societal benefits at scale in the physical world.
\end{IEEEbiography}
\vspace{-13pt}
\begin{IEEEbiography}[{\includegraphics[width=1in,height=1.25in,clip,keepaspectratio]{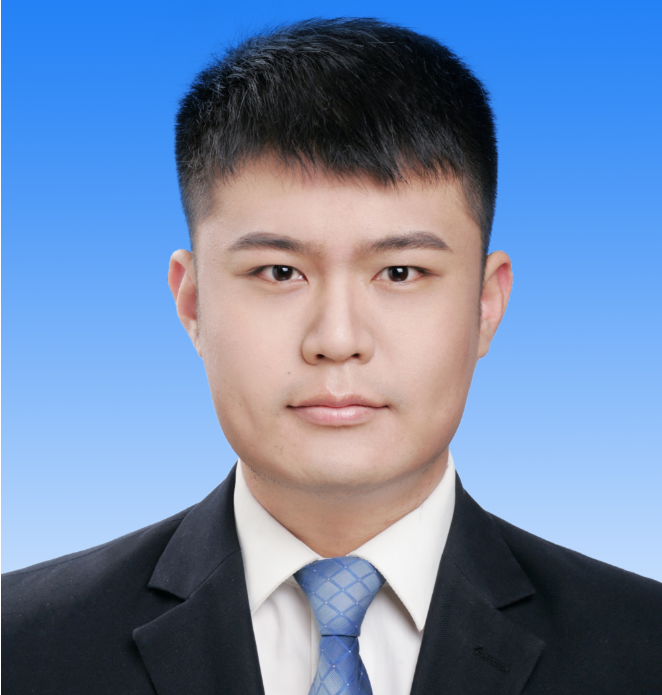}}]{Yao Mu} is an Assistant Professor at the Institute of Artificial Intelligence, Shanghai Jiao Tong University. He received his Ph.D. from the University of Hong Kong in 2025 and his Master's from Tsinghua University in 2021. His research in embodied intelligence, reinforcement learning, robot control, and autonomous driving has produced over 40 papers in top venues including RSS, NeurIPS, ICML, ICLR, and CVPR, accumulating more than 2500 citations. Dr. Yao's work has earned notable recognition including the ECCV Embodied Intelligence Workshop Best Paper Award, IEEE ICCAS 2020 Best Student Paper Award, and IEEE IV2021 Best Student Paper Nomination. He is a recipient of the Hong Kong Ph.D. Fellowship, University of Hong Kong Presidential Scholarship, and the National Scholarship of China. His research advances AI systems capable of effectively interacting with the physical world.
\end{IEEEbiography}

\begin{IEEEbiography}[{\includegraphics[width=1in,height=1.25in,clip,keepaspectratio]{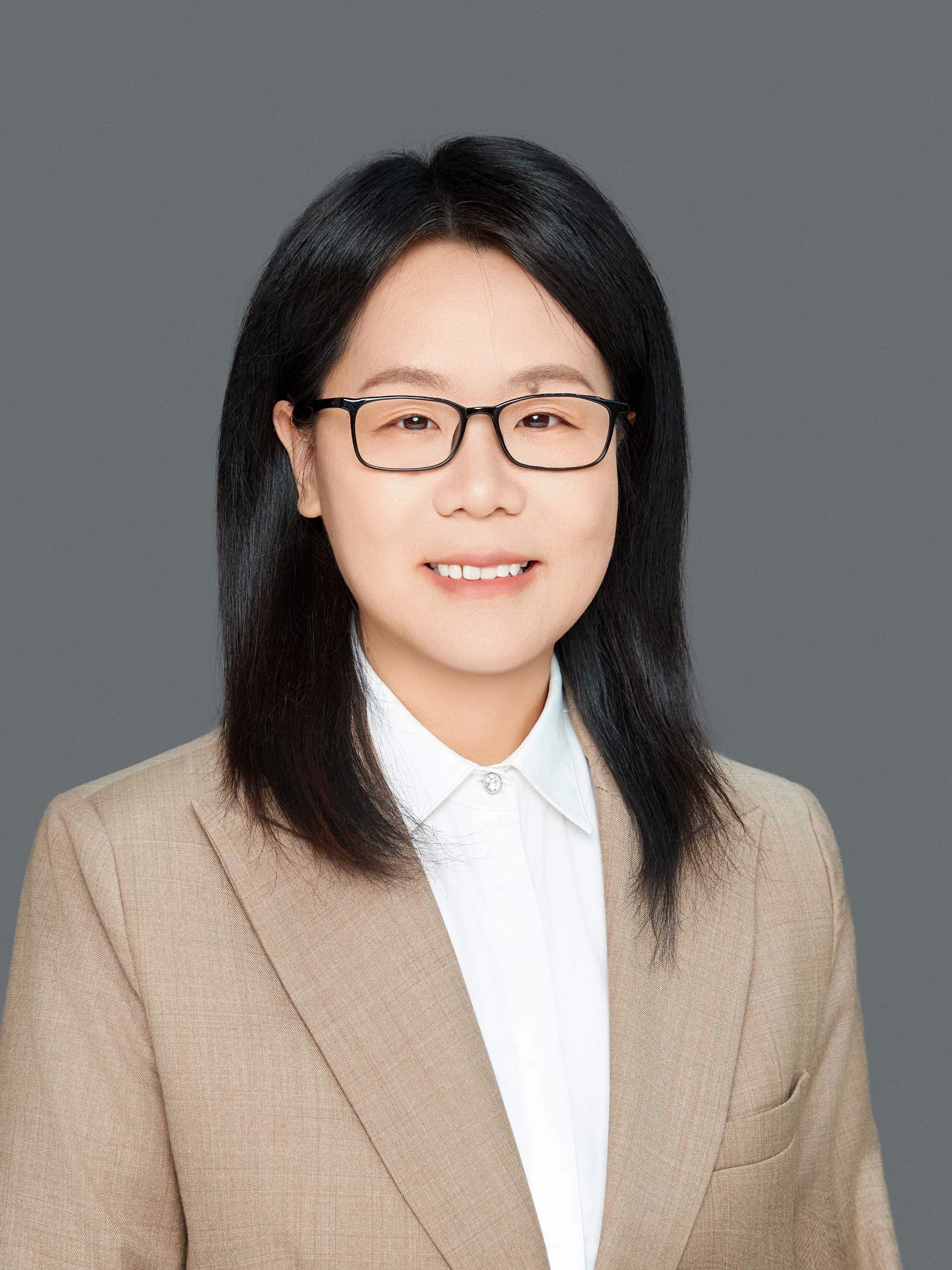}}]{Shufang Zhang} is currently an associate professor in the School of Electrical and Information Engineering, Tianjin University, Tianjin, China. She received her M.S. and Ph.D. degrees from Tianjin University in 2004 and 2007, respectively. Her research interests include Robot and SLAM, Artificial Intelligence, and Embodied Intelligence.
\end{IEEEbiography}
\vspace{-13pt}
\begin{IEEEbiography}[{\includegraphics[width=1in,height=1.25in,clip,keepaspectratio]{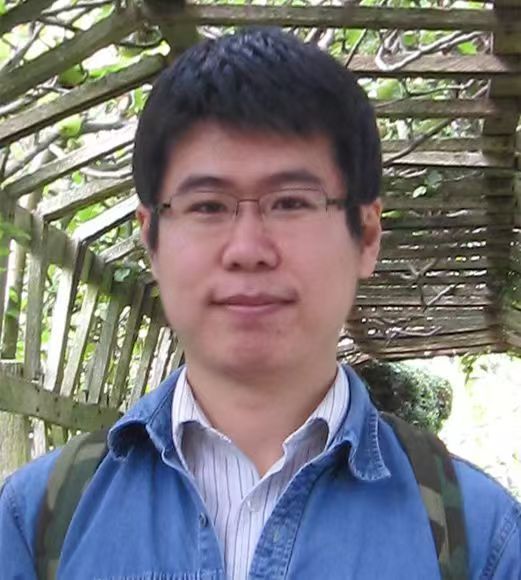}}]{Ran Song (Senior Member, IEEE)} is a Professor with the School of Control Science and Engineering, Shandong University, China since 2020. Before his current post, he was a senior lecturer at the University of Brighton, UK. He received his Ph.D. degree in electronic engineering from the University of York, UK in 2009 and his first degree from Shandong University in 2005. He has published more than 100 papers in peer-reviewed international conference proceedings and journals. His research interests lie in 3D visual perception, 3D vision for robotics, and robot learning.
\end{IEEEbiography}
\vspace{-13pt}
\begin{IEEEbiography}[{\includegraphics[width=1in,height=1.25in,clip,keepaspectratio]{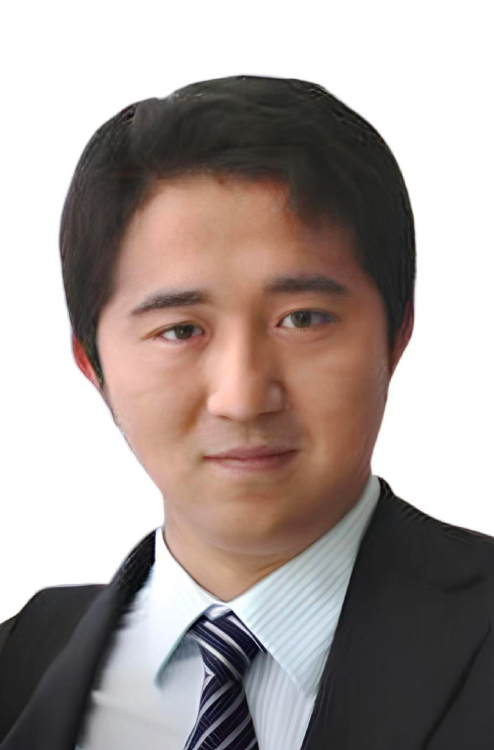}}]{Wei Zhang (Senior Member, IEEE)} received the Ph.D. degree in electronic engineering from the Chinese University of Hong Kong in 2010. He is currently a Professor with the School of Control Science and Engineering, Shandong University, Jinan, China. His research interests include computer vision and robotics. He has served as a program committee member and a reviewer for various international conferences and journals.
\end{IEEEbiography}
\vspace{-13pt}
\begin{IEEEbiography}[{\includegraphics[width=1in,height=1.25in,clip,keepaspectratio]{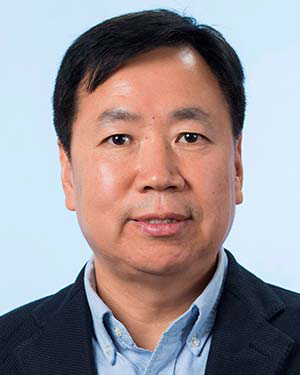}}]{Zeng-Guang Hou (Fellow, IEEE)} received the B.E. and M.E. degrees in electrical engineering from Yanshan University (formerly Northeast Heavy Machinery Institute), Qinhuangdao, China, in 1991 and 1993, respectively, and the Ph.D. degree in electrical engineering from the Beijing Institute of Technology, Beijing, China, in 1997. He is currently a Professor with the State Key Laboratory of Multimodal Artificial Intelligence
Systems, Institute of Automation, Chinese Academy of Sciences. His research interests include neural networks,
robotics, and intelligent systems. He is serving as a VP of the Asia Pacific Neural Network Society (APNNS) and Chinese Association of Automation (CAA). He is an associate editor of IEEE Transactions on Neural Networks and Learning Systems, IEEE Transactions on Cognitive and Developmental Systems, and Neural Networks, etc. He
is on the Board of Governors of International Neural Network Society (INNS). He was the Chair of Neural Network Technical Committee (NNTC) of Computational Intelligence Society (CIS), IEEE. Dr. Hou was a recipient of Neural Networks Best Paper Award in 2022, IEEE Transactions on Neural Networks Outstanding Paper Award in 2013.
\end{IEEEbiography}
\vspace{-13pt}
\begin{IEEEbiography}[{\includegraphics[width=1in,height=1.25in,clip,keepaspectratio]{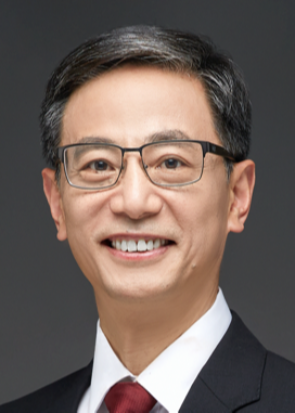}}]{Hong Zhang (Life Fellow, IEEE)} received his Ph.D. in Electrical Engineering from Purdue University in 1986. He was a Professor in the Department of Computing Science, University of Alberta, Canada, for over 30 years before he joined the Southern University of Science and Technology (SUSTech), China, in 2020, where he is currently a Chair Professor. Dr. Zhang served as the Editor-in-Chief of IROS Conference Paper Review Board (2020-2022) and as a member of the IEEE Robotics and Automation Society Administrative Committee (2023-25). He is a Life Fellow of IEEE and a Fellow of the Canadian Academy of Engineering. His research interests include robotics, computer vision, and image processing.
\end{IEEEbiography}







\end{document}